%% file: main.tex
\documentclass[10pt,twocolumn,letterpaper]{article}

\usepackage[pagenumbers]{cvpr} % To force page numbers, e.g. for an arXiv version

\usepackage{graphicx}
\usepackage{amsmath}
\usepackage{amssymb}
\usepackage{booktabs}
\usepackage[dvipsnames]{xcolor}

\usepackage[pagebackref,breaklinks,colorlinks]{hyperref}
\usepackage{amsmath}
\usepackage{siunitx}
\usepackage{tabularx}
\usepackage{float} % Added by Ted
\usepackage[accsupp]{axessibility}  % Improves PDF readability for those with disabilities.

\usepackage[capitalize]{cleveref}
\crefname{section}{Sec.}{Secs.}
\Crefname{section}{Section}{Sections}
\Crefname{table}{Table}{Tables}
\crefname{table}{Tab.}{Tabs.}

\newcolumntype{L}[1]{>{\raggedright\let\newline\\\arraybackslash\hspace{0pt}}m{#1}} % Added by Ted
\newcolumntype{R}[1]{>{\raggedleft\let\newline\\\arraybackslash\hspace{0pt}}m{#1}} % Added by Ted
\newcommand{\heading}[1]{\multicolumn{1}{c}{#1}} % Added by Ted
\newcommand*\samethanks[1][\value{footnote}]{\footnotemark[#1]} % Added by Ted
\def\cvprPaperID{2079} % *** Enter the CVPR Paper ID here
\def\confName{CVPR}
\def\confYear{2023}

\begin{document}

\title{SliceMatch: Geometry-guided Aggregation for Cross-View Pose Estimation}

\author{Ted Lentsch\thanks{\ indicates equal contribution.} \ \ \ \ \ \ \ \ \ \ Zimin Xia\samethanks \ \ \ \ \ \ \ \ \ \ Holger Caesar \ \ \ \ \ \ \ \ \ \ Julian F. P. Kooij \\
Intelligent Vehicles Group, Delft University of Technology, The Netherlands \\
{\tt\small \{T.deVriesLentsch,Z.Xia,H.Caesar,J.F.P.Kooij\}@tudelft.nl}
}
\maketitle

%%%%%%%%% ABSTRACT
\begin{abstract}
    This work addresses cross-view camera pose estimation, i.e., determining the 3-Degrees-of-Freedom camera pose of a given ground-level image w.r.t.~an aerial image of the local area.
    We propose SliceMatch, which consists of ground and aerial feature extractors, feature aggregators, and a pose predictor.
    The feature extractors extract dense features from the ground and aerial images.
    Given a set of candidate camera poses, the feature aggregators construct a single ground descriptor and a set of pose-dependent aerial descriptors.
    Notably, our novel aerial feature aggregator has a cross-view attention module for ground-view guided aerial feature selection and utilizes the geometric projection of the ground camera's viewing frustum on the aerial image to pool features.
    The efficient construction of aerial descriptors is achieved using precomputed masks.
    SliceMatch is trained using contrastive learning and pose estimation is formulated as a similarity comparison between the ground descriptor and the aerial descriptors.
    Compared to the state-of-the-art, SliceMatch achieves a 19\% lower median localization error on the VIGOR benchmark using the same VGG16 backbone at 150 frames per second,
    and a 50\% lower error when using a ResNet50 backbone.
    % Using the same backbone as the state-of-the-art, SliceMatch achieves a 19\% and 62\% lower median localization error on VIGOR and KITTI, and operates at more than 150 frames per second.
    
    % VGG-based performance
    % 19% = (6.25-5.07)/6.25, VIGOR same area pose estimation SliceMatch vs. MCC
    % 62% = (11.42-4.39)/11.42, KITTI same area pose estimation with 20 degrees prior, SliceMatch vs. LM

    % ResNet-based performance
    % Empty
    % Empty
\end{abstract}

%%%%%%%%% BODY TEXT
\input{1-introduction.tex}
\input{2-relatedwork.tex}
\input{3-methodology.tex}
\input{4-experiments.tex}
\input{5-conclusion.tex}
\section*{Acknowledgements}
% https://www.nwo.nl/financiering/hoe-werkt-dat/acknowledgement
This work is part of the research program Efficient Deep Learning (EDL) with project number P16-25, which is (partly) financed by the Dutch Research Council (NWO).

\clearpage % force placement of figures/tables before the references, so we get a feeling for the length of the paper while writing

%%%%%%%%% REFERENCES
{\small
\bibliographystyle{ieee_fullname}
\bibliography{main}
}

%%%%%%%%% APPENDIX
\newpage
\input{supp-overview}
\input{supp-text1}
\input{supp-text2}
\input{supp-text3}
\input{supp-text4}
\input{supp-text5}

\end{document}

%% file: 1-introduction.tex
\section{Introduction}\label{sec:introduction}
% \vspace{-2mm}

\begin{figure}[t]
    \centering    \includegraphics[width=0.945\columnwidth]{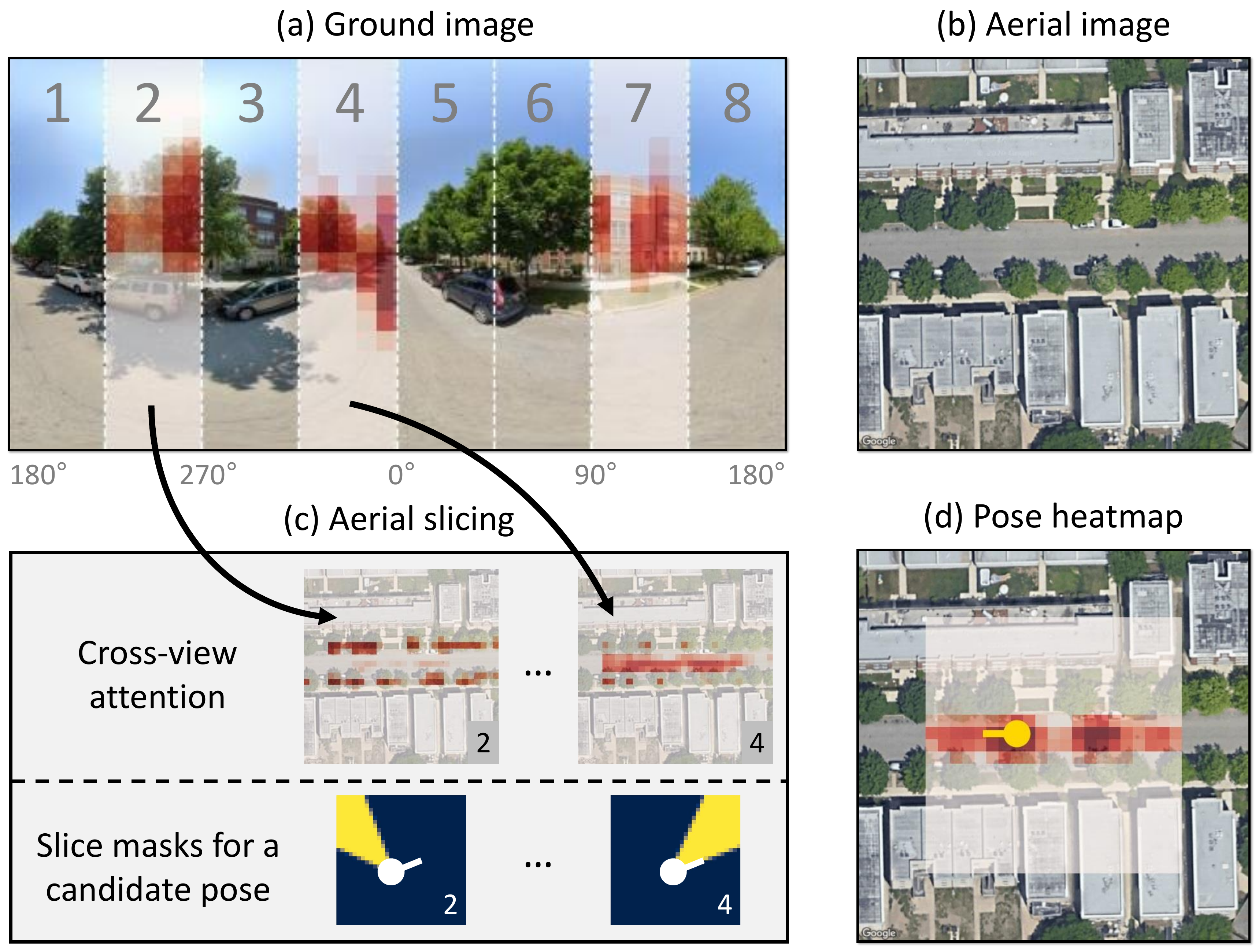}
    \caption{
    \textbf{SliceMatch identifies for a ground-level image (a) its camera's 3-DoF pose within a corresponding aerial image~(b).}
    It divides the camera's Horizontal Field-of-View (HFoV) into `slices', i.e., vertical regions in (a). After self-attention, our novel aggregation step (c) applies cross-view attention to create ground slice-specific aerial feature maps.
    % These are computed only once.
    To efficiently test many candidate poses, the slice features are aggregated using pose-dependent aerial slice masks that represent the camera's sliced HFoV at that pose. The slice masks for each pose are precomputed.
    % To efficiently test many candidate poses, for each pose, the slice features are aggregated using its precomputed aerial slice masks, representing the camera's sliced HFoV at that pose.
    % The slice masks are precomputed for each pose.
    All aerial pose descriptors are compared to the ground descriptor, resulting in a dense scoring map (d). Our output is the best-scoring pose.
    }
    \label{fig:introduction_image}
\end{figure}

Cross-view camera pose estimation aims to estimate the 3-Degrees-of-Freedom (3-DoF) ground camera pose, i.e., planar location and orientation, by comparing the captured ground-level image to a geo-referenced overhead aerial image containing the camera's local surroundings. % Degrees with "s" (Ted)
In practice, the local aerial image can be obtained from a reference database using any rough localization prior, e.g., Global Navigation Satellite Systems (GNSS), image retrieval~\cite{lowry2015visual}, or dead reckoning~\cite{king2006dead}.
However, this prior is not necessarily accurate, for example, GNSS can contain errors up to tens of meters in urban canyons~\cite{ben2011improving, zimin_gpsprior, zimin_gpsprior2}.
The cross-view formulation provides a promising alternative to ground-level camera pose estimation techniques that require detailed 3D point cloud maps~\cite{sattler2011fast} or semantic maps~\cite{cai2018integration, wang2021visual}, 
since the aerial imagery provides continuous coverage of the Earth's surface including the area where accurate point clouds are difficult to collect.
Moreover, acquiring up-to-date aerial imagery is less costly than maintaining and updating large-scale 3D point clouds or semantics maps.

Recently, several works have addressed cross-view camera localization~\cite{vigor} or 3-DoF pose estimation~\cite{zimin_metricloc, shi2022beyond, shi2022accurate, wang2022satellite}.
Roughly, those methods can be categorized into global image descriptor-based~\cite{vigor, zimin_metricloc} and dense pixel-level feature-based~\cite{shi2022beyond, shi2022accurate, wang2022satellite} methods.
Global descriptor-based methods take advantage of the compactness of the image representation and often have relatively fast inference time~\cite{vigor, zimin_metricloc}.
Dense pixel-level feature-based methods~\cite{shi2022beyond, shi2022accurate, wang2022satellite} are potentially more accurate as they preserve more details in the image representation.
They use the geometric relationship between the ground and aerial view to project features across views and estimate the camera pose via computationally expensive iterations.
Aiming for both accurate and efficient camera pose estimation, in this work, we improve the global descriptor-based approach and enforce feature locality in the descriptor.

We observe several limitations in existing global descriptor-based cross-view camera pose estimation methods~\cite{vigor, zimin_metricloc}.
First, they rely on the aerial encoder to encode all spatial context and the aerial encoder has to learn how to aggregate local information, e.g., via the SAFA module~\cite{safa}, into the global descriptor, without accessing the information in the ground view or exploiting geometric constraints between the ground-camera viewing frustum and the aerial image.
Second, existing global descriptor-based methods for cross-view localization~\cite{vigor,zimin_metricloc} do not explicitly consider the orientation of the ground camera in their descriptor construction.
As a result, they either do not estimate the orientation~\cite{vigor} or require multiple forward passes on different rotated samples to infer the orientation~\cite{zimin_metricloc}.
Third, existing global descriptors-based methods~\cite{vigor, zimin_metricloc} are not trained discriminatively against different orientations.
Therefore, the learned features may be less discriminative for orientation prediction.

To address the observed gaps, we devise a novel, accurate, and efficient method for cross-view camera pose estimation called SliceMatch (see Figure~\ref{fig:introduction_image}).
Its novel aerial feature aggregation explicitly encodes directional information and pools features using known camera geometry to aggregate the extracted aerial features into an aerial global descriptor.
The proposed aggregation step `slices' the ground Horizontal Field-of-View (HFoV) into orientation-specific descriptors.
For each pose in a set of candidates, it aggregates the extracted aerial features into corresponding aerial slice descriptors.
The aggregation uses cross-view attention to weigh aerial features w.r.t. to the ground descriptor, and exploits the geometric constraint that every vertical slice in the ground image corresponds to an azimuth range extruding from the projected ground camera position in the aerial image.
The feature extraction is done only once for constructing the descriptors for all pose candidates, resulting in fast training and inference speed.
We contrastively train the model by pairing the ground image descriptor with aerial descriptors at different locations and orientations.
Hence, the model learns to extract discriminative features for both localization and orientation estimation.

\textbf{Contributions:}
\textbf{i)}~A novel aerial feature aggregation step that uses a cross-view attention module for ground-view guided aerial feature selection, and the geometric relationship between the ground camera's viewing frustum and the aerial image to construct pose-dependent aerial descriptors.
\textbf{ii)}~SliceMatch's design allows for efficient implementation,
which runs significantly faster than previous state-of-the-art methods.
Namely, for an input ground-aerial image pair, SliceMatch extracts dense features only once, aggregates aerial descriptors at a set of poses without extra computation, and compares the aerial descriptor of each pose with the ground descriptor.
\textbf{iii)}~Compared to the previous state-of-the-art global descriptor-based cross-view camera pose estimation method,
SliceMatch constructs orientation-aware descriptors and adopts contrastive learning for both locations and orientations.
Powered by the above designs, SliceMatch sets the new state-of-the-art for cross-view pose estimation on two commonly used benchmarks.

%% file: 2-relatedwork.tex
\section{Related Work}\label{sec:related_work}

\noindent Here, we review the work most related to SliceMatch.

\textbf{Cross-view image retrieval} is the task of finding matching aerial image patches from a reference database for a query ground image. The location of the retrieved aerial patch can be used as a localization estimate for the query image~\cite{workman2015location, lin2015learning, cvusapaperoriginal}. 
In general, this task is done by creating a global image descriptor for the query ground image and each reference aerial patch~\cite{cvmnet, safa, liu2019lending, revisiting_orientation, shi2020optimal, toker_gan, cvm_transformer, vigor, lu2022s}.
Different approaches have been proposed to build discriminative descriptors. CVM-Net~\cite{cvmnet} uses NetVLAD~\cite{netvlad} to build viewpoint invariant descriptors.
In~\cite{safa}, spatial attention modules are used to extract corresponding features across views.
L2TLR~\cite{cvm_transformer} exploits the positional encoding of Transformers~\cite{vaswani2017attention} to learn geometric correspondences between the ground and aerial view.
TransGeo~\cite{zhu2022transgeo} uses attention-guided non-uniform cropping to only pay attention to informative regions in Transformers.
Apart from advanced architectures, several works~\cite{safa, li2022multi, regmi_gan, toker_gan, shi2022accurate} have tried to synthesize one view using another to bridge the domain gap between the ground and aerial images.
Besides, some works~\cite{safa, shi2022cvlnet, wang2021each} use the geometric relationship between vertical lines in the ground image and azimuth directions in the aerial image to ease the learning or to estimate the orientation of the ground camera~\cite{shi2020looking, revisiting_orientation, shi2022accurate}.
A few works have tried to explicitly enforce feature locality in global representations~\cite{rodrigues2022global, wang2021each}, but they assume that the camera is located at the center of the aerial image.
This limits the generalization of these methods to pose estimation.

\textbf{Cross-view camera pose estimation} works~\cite{graphbased, vigor, zimin_metricloc, shi2022beyond, shi2022accurate} go a step further than retrieval and aim to determine the location and orientation of the ground camera in the matching aerial image. 
A landmark graph matching-based method is used in \cite{graphbased}, but a separate object detector is needed.  
In~\cite{cvusapaper}, the semantic segmentation of the ground-level image is compared to the ground-level semantic map predicted from the aerial image for estimating the location and orientation of the ground camera.
Recently, \cite{vigor} proposes a model that first retrieves an aerial image for a query image and then uses a multilayer perceptron to regress the query's location using the global image descriptors. Later, \cite{zimin_metricloc} formulates the localization problem as a multi-class classification problem and their model produces a dense multi-modal distribution to capture localization ambiguity. \cite{shi2022beyond} projects the dense aerial features to ground perspective view based on homography and iteratively estimates the ground camera pose using Levenberg–Marquardt algorithm~\cite{levenberg1944method, marquardt1963algorithm}.
However, the iterative process is computationally expensive (e.g. $\sim2$ frames per second~\cite{shi2022beyond}) and it requires an accurate initial estimate to converge to a
good local optimum.
In~\cite{wang2022satellite}, the ground image is fused with LiDAR data for iterative pose estimation. \cite{shi2022accurate} samples aerial patches in the retrieved aerial image and applies a projective transformation on each sampled patch. Localization is achieved by selecting the location of the patch with the highest similarity to the ground image in the feature space. However, the computation increases linearly with the number of sampled locations.
Lastly,~\cite{hu2022beyond} estimates only the orientation using the known location of the ground camera in the aerial image.
So far, existing end-to-end methods are either global image descriptor-based~\cite{vigor, zimin_metricloc} or dense local feature-based~\cite{shi2022beyond, wang2022satellite, shi2022accurate, hu2022beyond}.
We argue that enforcing the right amount of feature locality in global image descriptors can be a promising direction toward the accuracy and runtime requirements of autonomous driving~\cite{reid2019localization}.

\textbf{Bridging ground and aerial views} is relevant in many other research directions.
For example, ground-to-Bird's-Eye-View (BEV) semantic mapping~\cite{roddick2020predicting, philion2020lift, saha2021translating} tries to map the semantics in the ground perspective view to BEV given the known ground camera pose and intrinsics. 
\cite{roddick2020predicting} introduces a dense Transformer layer to condense the ground image features along the vertical dimension and then predicts features along the depth axis in a polar coordinate system. \cite{philion2020lift} lifts the ground features along the depth dimension and then projects them to BEV.
In \cite{saha2021translating}, a Transformer is used for the ground image column-to-BEV polar ray mapping. 
A few works~\cite{lu2020geometry,shi2022geometry} synthesize ground-level panoramas from aerial images.
All aforementioned works utilize the geometric relationship between the ground-level camera's frustum and BEV.
Indoor localization using floor maps~\cite{howard2021lalaloc, min2022laser, howard2022lalaloc++} is also a relevant research direction. 
LaLaLoc~\cite{howard2021lalaloc} renders the ground view using the 2D floor plan and optimizes the ground camera pose estimation using the global representation of the ground-level query image and the global representation of the rendered view.
LaLaLoc++~\cite{howard2022lalaloc++} removes the
need for explicit modeling or rendering in \cite{howard2021lalaloc} by introducing a global floor plan comprehension module.
LASER~\cite{min2022laser} constructs a geometrically-structured latent space by aggregating viewing ray features for Monte Carlo Localization in 2D point cloud floor maps.
It needs the occupancy boundaries information (e.g. walls) to form the 2D point cloud input. Thus it is not directly generalizable to aerial imagery.

%% file: 3-methodology.tex
\section{Methodology}\label{sec:methodology}

We explain the cross-view camera pose estimation task, our SliceMatch method, and its novel aggregation step.
% and the used training loss.

\subsection{Cross-View Camera Pose Estimation}

Given a ground-level image $I_g$ and a square overhead aerial image $I_a$ that contains the local surroundings of $I_g$, we aim to determine the 3-DoF pose, $\xi=(u,v,\theta)$, of the ground camera that captured $I_g$.
Here, $(u,v) \in [0,1]^2$ are the image coordinates in $I_a$, and $\theta \in [0,360^{\circ})$ is the camera orientation, i.e., the angle from the North direction clockwise to the center line (the `front' direction) of the ground camera projected onto the aerial view.
Ground images can either be panoramic or have a limited HFoV.
Similar to~\cite{shi2022beyond}, we assume that the ground camera's pitch and roll are small.

\subsection{SliceMatch Overview}
\begin{figure*}[t]
    \centering
    \includegraphics[height=6.4cm]{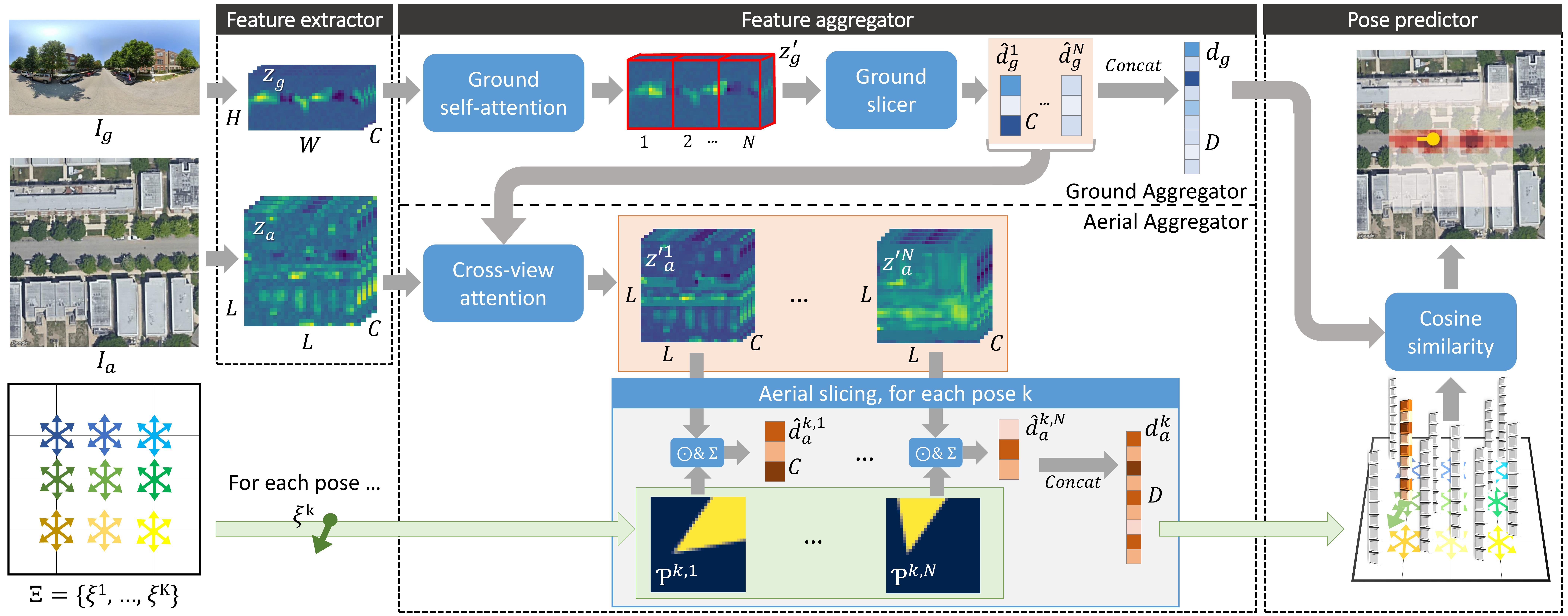}
   \caption{
    \textbf{The SliceMatch pipeline.}
    The input to SliceMatch is a pair a ground-aerial image pair and a set of K candidate ground camera poses. 
    SliceMatch consists of ground and aerial feature extractors, feature aggregators, and a pose predictor.
    In the shown output image, we overlayed the matching scores for all poses on the input aerial image.
    The predicted pose is the one with highest score.
    }
    \label{fig:methodology_slicematch_architecture}
\end{figure*}

SliceMatch explicitly separates feature extraction and aggregation, where the latter exploits geometric knowledge on how the ground camera's viewing frustum projects on the aerial image.
In SliceMatch, pose estimation is formulated as an efficient process that compares aerial descriptors for a set $\Xi = \{ \xi^{1}, \cdots, \xi^{K} \}$ of $K$ candidate poses to the ground image descriptor.
During training, the set consists of $K_{train}$ poses at a fixed uniform grid in 3-DoF pose space.
During inference, we use $K_{test}$ poses ($K_{test}>K_{train}$), and the predicted pose is the candidate for which its aerial descriptor is most similar to the ground descriptor.
See Figure~\ref{fig:methodology_slicematch_architecture} for an overview of the method.
We discuss each step next.

% MACROS
%\newcommand{\sliceDim}{\widehat{D}}
\newcommand{\sliceDim}{C}
\newcommand{\globalDim}{D}
\newcommand{\featmapA}{z_a}
\newcommand{\featmapG}{z_g}
\newcommand{\featmapWeightedA}{z_a'}
\newcommand{\featmapWeightedG}{z_g'}

\newcommand{\sliceFeatSetG}{\widehat{\Delta}_g}
\newcommand{\sliceFeatSetA}{\widehat{\Delta}_a}
\newcommand{\globalFeatSetA}{\Delta_a}

\newcommand{\sliceFeatG}{\hat{d}_g}
\newcommand{\sliceFeatA}{\hat{d}_a}
\newcommand{\globalFeatG}{d_g}
\newcommand{\globalFeatA}{d_a}

\textbf{Feature extractor:}
Input images $I_g$ and $I_a$ are first mapped to feature maps, $\featmapG=f_g(I_g) \in \mathbb{R}^{H \times W \times \sliceDim}$ and $\featmapA=f_a(I_a) \in \mathbb{R}^{L \times L \times \sliceDim}$,
where $f_g$ and $f_a$ can be any convolutional backbone (e.g. VGG~\cite{vgg16} or ResNet~\cite{he2016deep}).
We adopt the commonly used setup that $f_g$ and $f_a$ have the same architecture without weight-sharing~\cite{vigor, zimin_metricloc}.
We seek translational equivariance in our encoders, and thus do not focus on Vision Transformers~\cite{dosovitskiy2020image} in this work.

\textbf{Feature aggregator:}
Our novel aggregator step efficiently constructs a single ground and multiple pose-dependent aerial descriptors from the extracted image features
through the use of `slices'.
In our work, each slice represents a non-overlapping range in the azimuth viewing direction,
and is used to aggregate the local image features within that azimuth range. 
In the ground view, a slice thus corresponds to a vertical rectangular region in the image/feature map,
and in the aerial view, it is a triangle-shaped region extending from a candidate pose (see Figure~\ref{fig:introduction_image}).
This will be explained in more detail in Section~\ref{sec:feature-aggregation}.
We refer to an aggregated feature in a single slice as a ground/aerial \textit{slice descriptor},
containing the visual information for that viewing direction.
Likewise, we refer to a ground/aerial \textit{global descriptor} as the concatenation of the slice descriptors of all pose-relative orientations, representing the full HFoV of the ground camera.

Concretely,
the extracted feature maps $\featmapG$ and $\featmapA$
are fed into our heterogeneous feature aggregators,
as shown in Figure~\ref{fig:methodology_slicematch_architecture}.
The ground aggregator $\textit{agg}_g(\featmapG)$ generates a
set\footnote{
Note that we use $\hat{d}$ and $\hat{\Delta}$ (with hat) to indicate \textit{slice} descriptors/sets,
and $d$ and $\Delta$
(without hat) to indicate \textit{global} descriptors/sets.
}
$\sliceFeatSetG = \{ \sliceFeatG^1, \cdots, \sliceFeatG^N \}$ of $\sliceDim$-dimensional slice descriptors $\sliceFeatG^n$ for $N$ azimuth directions, where $N$ is a hyperparameter for the number of slices.
The ground global descriptor $\globalFeatG = \textit{Concat}(\sliceFeatG^1, \cdots, \sliceFeatG^N)$
is thus a vector of length $\globalDim = N \cdot \sliceDim$.
The aerial aggregator $\textit{agg}_a(\featmapA, \sliceFeatSetG, \Xi)$ receives the aerial features $\featmapA$, the ground slice descriptors $\sliceFeatSetG$, and the set of $K$ poses $\Xi$.
It generates $\globalFeatSetA = \{\globalFeatA^1, \cdots, \globalFeatA^K \}$, the set of $K$ pose-dependent aerial global descriptors $\globalFeatA^k \in \mathbb{R}^D$.
Section~\ref{sec:feature-aggregation}
will discuss both aggregators in detail.

\textbf{Pose predictor:}
The pose predictor receives the ground global descriptor $\globalFeatG$, and
the set $\globalFeatSetA$ that contains the $K$ aerial global descriptors corresponding to the candidate poses in set $\Xi$.
We compute the cosine similarity $c^k$ between $\globalFeatG$ and all $\globalFeatA^k \in \globalFeatSetA$ and,
during inference, use $\xi^k$ corresponding to the highest similarity value $c^k_{max} = {max}(c^1, \cdots, c^K)$ as the predicted pose.
Note that similar to~\cite{zimin_metricloc}, we obtain a heatmap that can express multimodal pose estimation ambiguity, which can be beneficial for downstream fusion.

\textbf{Loss Function:}
We modify the infoNCE loss~\cite{oord2018representation} from contrastive representation learning~\cite{khosla2020supervised} to train SliceMatch.
Using $K = K_{train}$ training poses, our loss $\mathcal{L}$ is defined as,
\begin{equation}
\begin{aligned}
\mathcal{L} = - \log  \left( \frac{\exp (c^{GT} / \tau)}{\frac{\alpha}{K}  \sum_{k=1}^{K} \exp (c^k / \tau) + \exp (c^{GT}/\tau)} \right).
\end{aligned}
\label{eq:methodology_infonce_loss}
\end{equation}

In Equation~\eqref{eq:methodology_infonce_loss}, $\alpha$ is our introduced hyperparameter that weighs the contribution of $K$ poses to the learning. Variable $c^{GT}$ is the cosine similarity between $\globalFeatG$ and $\globalFeatA^{GT}$ at $\xi^{GT}$, and $c^k$ is that between $\globalFeatG$ and $\globalFeatA^k$ at $\xi^k$. Hyperparameter~$\tau$ is proposed in \cite{oord2018representation}.
The original infoNCE loss in~\cite{oord2018representation} can be acquired using $\alpha=K$.
With $\mathcal{L}$, we contrast the ground truth pose with $K_{train}$ other poses at different locations and orientations, thus the model learns to extract discriminative features for both location and orientation prediction.

\subsection{Geometry-Guided Cross-View Aggregation}
\label{sec:feature-aggregation}
Here, we describe the novel aggregation step in more detail.
Unlike the SAFA module~\cite{safa} used in \cite{vigor, zimin_metricloc},
our aggregation uses geometric knowledge on how the views should spatially relate.
Ground-to-aerial attention further improves quality, 
as the visual information in each ground slice
informs what aerial features are relevant to produce the corresponding aerial slice descriptors,
thus promoting shared features specific to each viewing direction.

\subsubsection{Ground Feature Aggregator}
To summarize the important features in each vertical slice in the ground camera's viewing frustum, we construct our ground feature aggregator $\textit{agg}_g(\featmapG)$ with a self-attention module and a feature slicer.
Since not all information in ground image $I_g$ will be present in the aerial image $I_a$ (e.g. sky and transient objects),
the self-attention module re-weighs $\featmapG$ along the spatial dimensions $H$ and $W$,
\begin{gather}
    \featmapWeightedG = \mathcal{M}_g \odot \featmapG, \quad \mathcal{M}_g = \textit{Sigmoid}(\textit{Conv}_{1\times1}(\featmapG))   \label{eq:ground_mask}.
\end{gather}

Here, $\mathcal{M}_g$ is a learned mask with shape $H \times W \times 1$ that re-weighs the ground feature map $\featmapG$ into $\featmapWeightedG$.
The \textit{Sigmoid} operation enforces the weights in $\mathcal{M}_g$ are between $0$ and $1$.
The $\odot$ denotes element-wise multiplication, with the ability to broadcast the mask $\mathcal{M}_g$ over all channels of $\featmapG$.

The ground slicer then divides $\featmapWeightedG$ into $N$ vertical slices,
cutting the feature map along the horizontal (azimuth) direction.
For each slice, a normalized slice descriptor is computed by averaging all features within the slice and applying L2 normalization.
This results in the set $\sliceFeatSetG = \{\sliceFeatG^1, 
\cdots, \sliceFeatG^N \}$ of N ground slice descriptors.
Each slice local descriptor thus represents the model's attended feature in the corresponding vertical slice (i.e. an azimuth range) in the ground camera's viewing frustum.
The ground global descriptor is obtained by concatenating all $N$ ground slice descriptors, i.e. $\globalFeatG = \textit{Concat}(\sliceFeatG^1, \cdots, \sliceFeatG^N)$.

\subsubsection{Aerial Feature Aggregator}
The aerial aggregator $\textit{agg}_a(\featmapA, \sliceFeatSetG, \Xi)$ has a similar role as the ground aggregator, but its feature selection is also conditioned on the ground slice descriptors $\sliceFeatSetG$ using a cross-view attention module and the set of poses $\Xi$ for geometry-guided feature aggregation.

\textbf{Cross-view attention:}
Since in the ground view most content that is seen in the aerial view will be occluded, we propose a cross-view attention module to specifically extract the aerial features that should match the visible content of each ground slice.
In detail, we match the $\sliceDim$-dimensional aerial feature $\featmapA^{i,j}$ at each spatial location $(i,j)$ with $1 \leq i \leq L, 1 \leq j \leq L$ in the aerial feature map $\featmapA$ to each ground slice descriptor $\sliceFeatG^n \in \sliceFeatSetG$ to acquire a similarity score map $S^n$ of size $L \times L$, where $S^{n,i,j} = \textit{Sim}(\sliceFeatG^n, \featmapA^{i,j})$.
In total, there are $N$ similarity score maps, i.e. one for each ground slice descriptor.
Then, we treat each $S^n$ as extra features 
and concatenate it along the feature dimension with aerial feature map $\featmapA$ 
\cite{zimin_metricloc},
and use these extended features to produce a cross-view attention mask,
\begin{align}
  \mathcal{M}_a^n = \textit{Sigmoid}(\textit{Conv}_{1\times1}(\textit{Concat}(\featmapA, S^n))).
  \label{eq:crossview_attention}
\end{align}

We thus get in total $N$ cross-view masks $\mathcal{M}_a^n$.
Each of these denotes the importance of the aerial features w.r.t.~the $n$-th ground slice descriptor $\sliceFeatG^n$.
Finally, we re-weigh $\featmapA$ for each ground slice descriptor,
giving us $N$ re-weighted aerial feature maps $\featmapWeightedA^n$ of size $L \times L \times \sliceDim$,
i.e.~$\featmapWeightedA^n = \mathcal{M}_a^n \odot \featmapA$.

\textbf{Geometry-guided feature aggregation:}
Finally, the $K$ pose-dependent aerial descriptors $\globalFeatA^k$
can be constructed for the candidate poses in $\Xi$.
For each pose $\xi^k$, we can precompute $N$ aerial slice masks $\mathcal{P}^{k,n} \in [0, 1]^{L \times L}$, $1 \leq n \leq N$.
The slice mask $\mathcal{P}^{k,n}$ expresses the geometry of the ground camera's viewing frustum in the aerial feature map for the $n$-th orientation slice, assuming that the camera would have the $k$-th pose.
Each cell in the slice mask contains a value in the range $[0,1]$ proportional to how much of that cell intersects this frustum, so $1.0$ for fully contained cells, $0.0$ for cells fully outside the frustum, and an intermediate value for cells that partially overlap.

With the slice masks, the $n$-th aerial slice descriptor at pose $k$ can be computed efficiently.
For each of the $\sliceDim$ channels,
we compute a weighted average over all of the $L \times L$ spatial locations $(i,j)$ in the feature map $\featmapWeightedA^n$,
using the elements of slice mask $\mathcal{P}^{k,n}$ as weights. After L2~normalization,
we obtain aerial slice
descriptor $\sliceFeatA^{k,n}$,
\begin{align}
\sliceFeatA^{k,n} &= \textit{Norm} \Bigl(
\frac{1}{\sum_{i,j} \mathcal{P}^{k,n}_{i,j}}
\sum_{i,j} \left( \mathcal{P}^{k,n} \odot \featmapWeightedA^n \right)_{i,j} \Bigr).
\label{eq:aerial_slice}
\end{align}

Analogous to the ground view, the $k$-th pose's global descriptor is obtained using
$\globalFeatA^k = \textit{Concat}(\sliceFeatA^{k,1}, \cdots, \sliceFeatA^{k,N})$.

\textbf{Efficient implementation:}
A benefit of our proposed architecture is that the computational complexity of most operations is independent of the number of candidate poses $K$.
The main cost to increase $K$,
and therefore improve accuracy by testing more diverse poses at inference time,
is to add more precomputed slice masks, and perform the additional multiplications and normalizations for Equation~\eqref{eq:aerial_slice} and the final cosine similarity comparison.
These are simple operations that can be highly optimized and parallelized in the implementation, and we will show that testing more candidate poses does not increase our runtime.

%% file: 4-experiments.tex
\section{Experiments}\label{sec:experiments}

We first introduce the used datasets and the evaluation metrics. After that, our implementation details and ablation studies are presented. Finally we quantitatively and qualitatively compare SliceMatch to state-of-the-art baselines.

\begin{figure*}[t]
    \centering
    \subfloat{{\includegraphics[width=3.3cm]{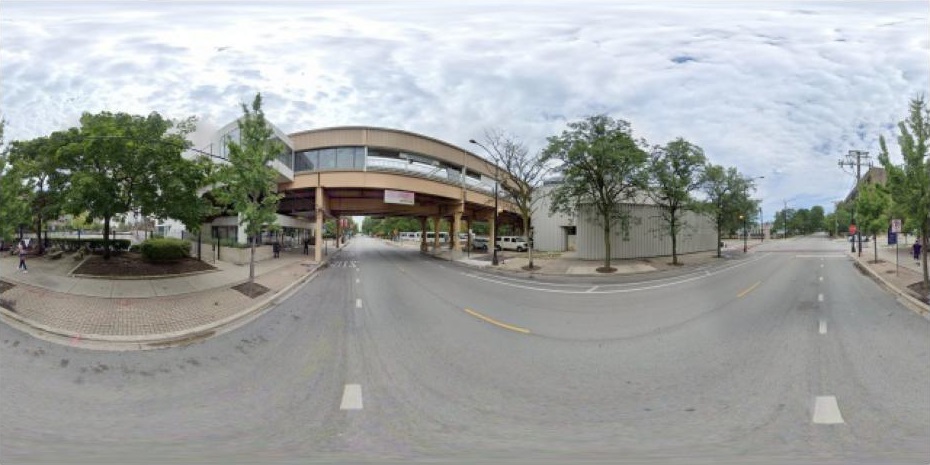}}}
    \hspace{-5mm}%
    \qquad
    \subfloat{{\includegraphics[width=3.3cm]{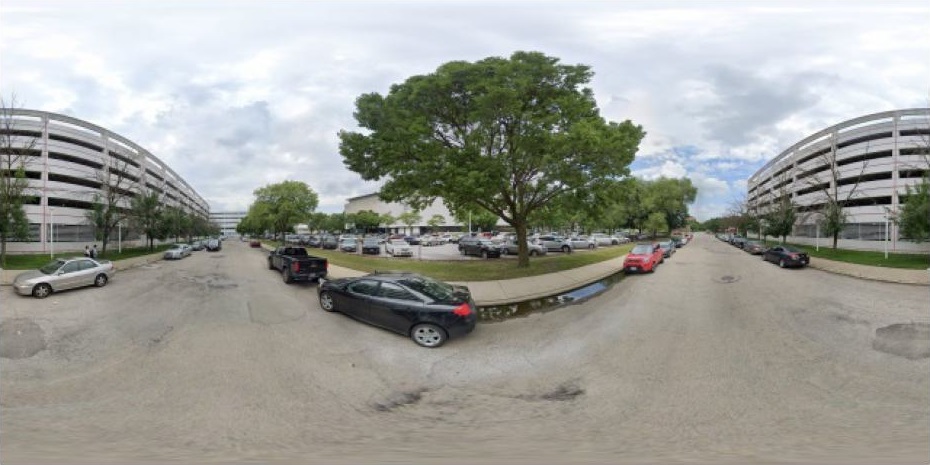}}}
    \hspace{-5mm}%
    \qquad
    \subfloat{{\includegraphics[width=3.3cm]{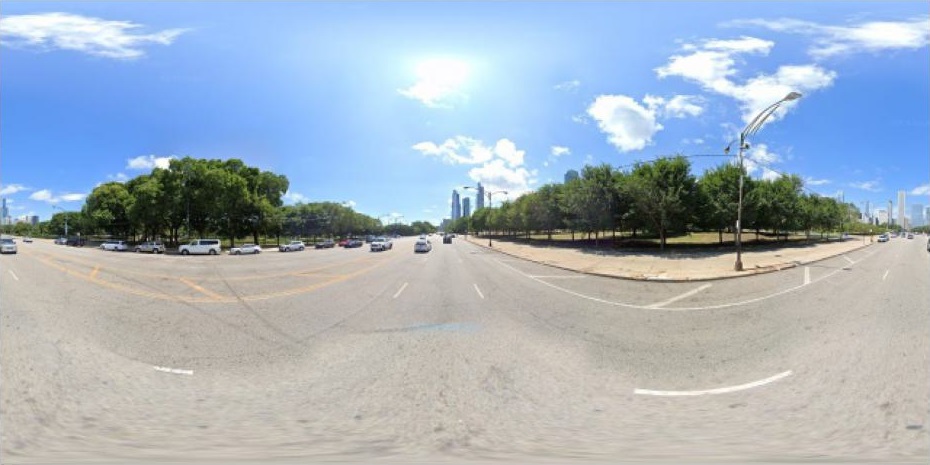}}}
    \hspace{-5mm}%
    \qquad
    \subfloat{{\includegraphics[width=3.3cm]{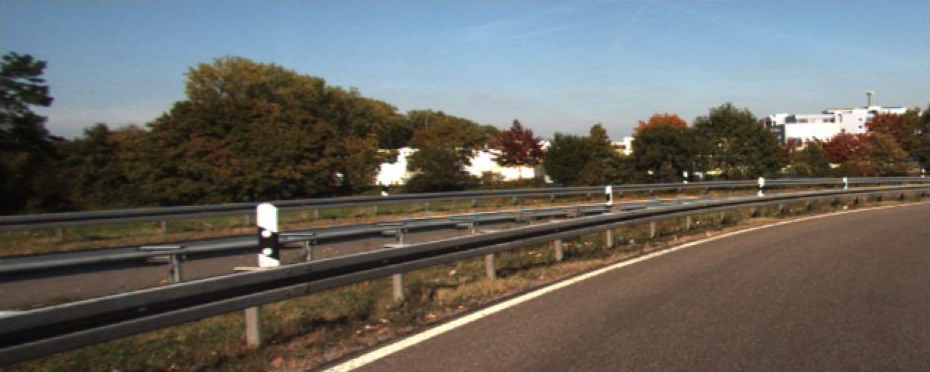}}}
    \hspace{-5mm}%
    \qquad
    \subfloat{{\includegraphics[width=3.3cm]{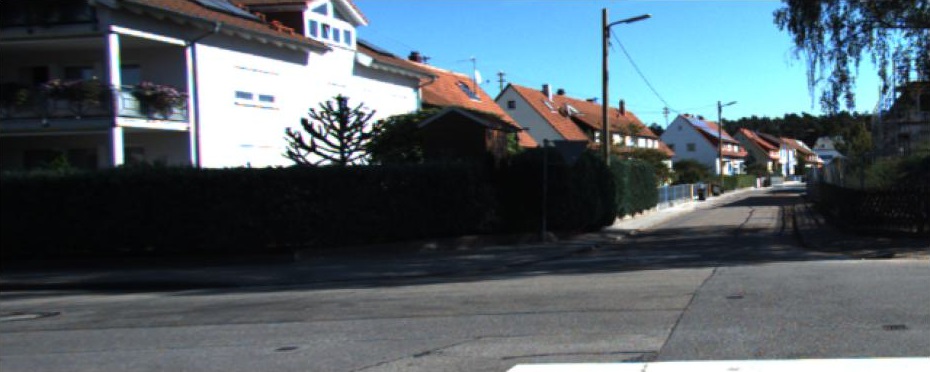}}}
    \vspace{2mm}%
    \qquad
    \subfloat[\centering VIGOR example 1]{\setcounter{subfigure}{0}%resets subfigure to (a)
    {\includegraphics[width=3.3cm]{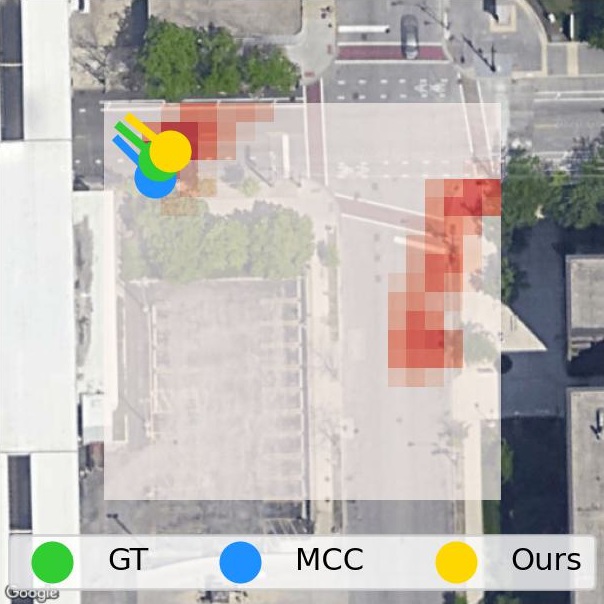}}}
    \hspace{-5mm}%
    \qquad
    \subfloat[\centering VIGOR example 2]{{\includegraphics[width=3.3cm]{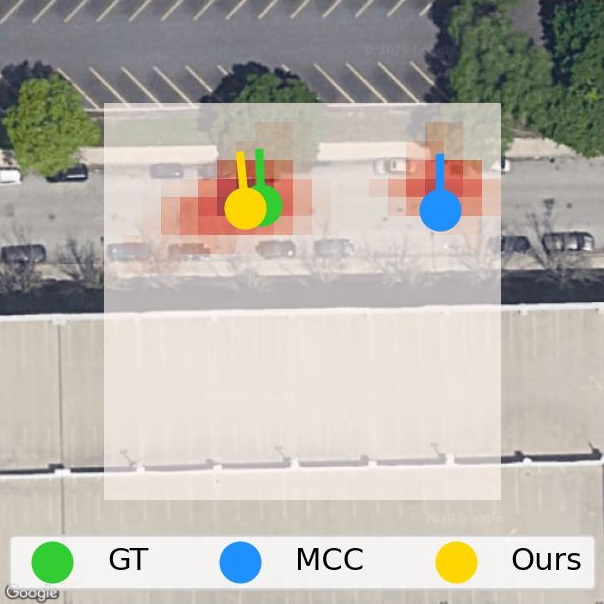}}}
    \hspace{-5mm}%
    \qquad
    \subfloat[\centering VIGOR example 3]{{\includegraphics[width=3.3cm]{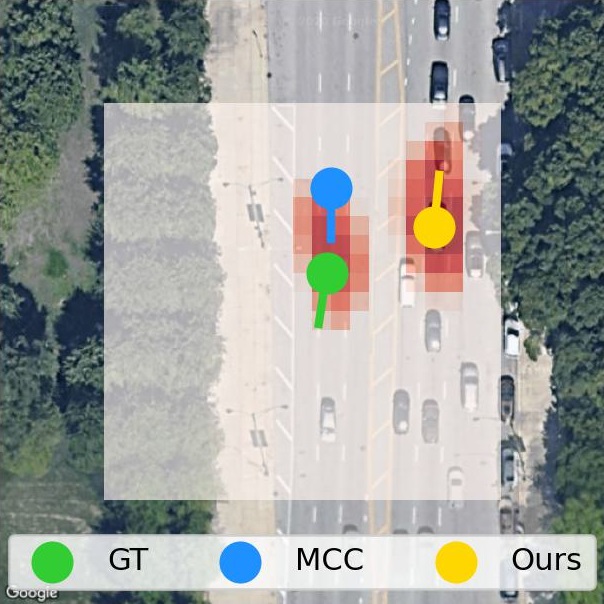}}\label{fig:failure_case}}
    \hspace{-5mm}%
    \qquad
    \subfloat[\centering KITTI example 1]{{\includegraphics[width=3.3cm]{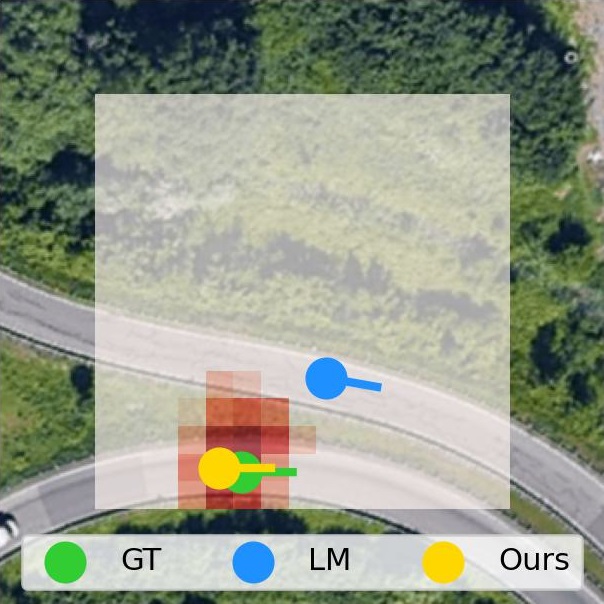}}}
    \hspace{-5mm}%
    \qquad
    \subfloat[\centering KITTI example 2]{{\includegraphics[width=3.3cm]{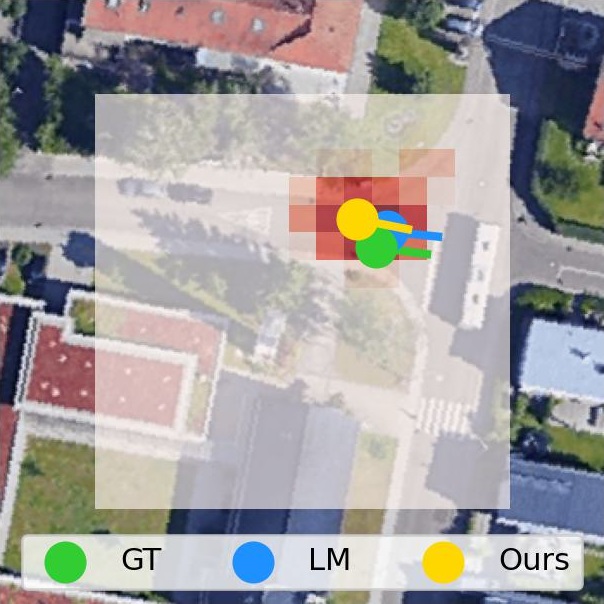}}}
    \caption{
    \textbf{Qualitative evaluation of SliceMatch on VIGOR~\cite{vigor} and KITTI~\cite{geiger2013vision, shi2022beyond}.}
    %VIGOR: (a), (b), (c). KITTI: (e), (f).
    Top row: input ground image. Bottom row: GT and pose estimation results overlayed on input aerial image.
    Red shading indicates highest similarity score between the ground descriptor and the aerial descriptors among all orientations at that location.
    (c) shows a SliceMatch failure: the best match is in the wrong mode.
    }
    \vspace{-3mm}
    \label{fig:slicematch_visuals}
\end{figure*}

\subsection{Datasets}\label{subsec:datasets}

\textbf{VIGOR dataset}~\cite{vigor} contains geo-tagged ground-level panoramas and aerial images collected in 4 cities in the US. As defined in~\cite{vigor}, each ground panorama has 1 positive and 3 semi-positive aerial images.
An aerial image is positive if the ground camera's location is within the aerial image's center quarter area, otherwise, it is semi-positive. 
Importantly, we found that the original ground truth locations in~\cite{vigor} can contain errors up to 3 meters due to the use of wrong ground resolutions (0.114m/pixel) of the aerial images, thus we created and use here corrected labels, see details in Supplementary Material.
For training and testing our method and baselines, we use positive aerial images and corrected ground truth
(we reran all baselines since quantitative results with the new labels differ slightly from those reported in the literature).
We adopt the same-area and cross-area splits from \cite{vigor} to test the model’s generalization to new measurements in the same cities and across different cities. Besides, we use the same-area training dataset of New York as a tuning split for the ablation study.

\textbf{KITTI dataset}~\cite{geiger2013vision} contains ground-level images with a limited HFoV taken by a moving vehicle from different trajectories at different times and \cite{shi2022beyond} augmented the dataset with aerial images. We use their split.
The Training and Test1 sets are different measurements from the same region, while the Test2 set has been captured in a different region.

\subsection{Evaluation Metrics}

We follow the convention of~\cite{zimin_metricloc} and report the mean and median error in meters between the predicted and ground truth location over all test image pairs.
Similarly, for orientation prediction, we report the mean and median absolute angular difference between the predicted and ground truth orientation in degrees.
Following \cite{shi2022beyond}, for the KITTI dataset, we additionally include the recall under a certain threshold for longitudinal (driving direction) and lateral localization error, and orientation estimation error.
Our thresholds are set to 1m and 5m for localization and to 1\textdegree{} and 5\textdegree{} for orientation estimation.

\subsection{Implementation Details}

As in~\cite{vigor,zimin_metricloc}, we use VGG16~\cite{vgg16} up to stage 5 for the feature extractors $f_g$ and $f_a$.
The pooling operation of the last layer is removed. 
The spatial size of $I_g$ is $320 \times 640$ on VIGOR dataset and $256 \times 1024$  on KITTI dataset, and that of $I_a$ is $512 \times 512$ on both datasets.
This results in feature maps with $H \times W = 20 \times 40$ / $16 \times 64$ on VIGOR / KITTI, $L \times L = 32 \times 32$, and $C=512$.
The feature extractors do not share their weights and are pre-trained on ImageNet~\cite{imagenet_dataset}.
In Equation~\eqref{eq:ground_mask} and~\eqref{eq:crossview_attention}, $\textit{Conv}_{1\times1}$ consists of two sequential convolution layers with a kernel size of 1 and a ReLU activation in between.
During training, SliceMatch is trained end-to-end using Adam optimizer~\cite{kingma2014adam} with a learning rate of \num{1e-5}, and we use a batch size of 4.
To get a set of candidate camera poses $\Xi$, we use poses at a uniform grid of $7 \times 7$ locations $\times~16$ orientations on VIGOR, and $5 \times 5$ locations $\times~16$ orientations on KITTI during training.
For inference, we use $21 \times 21 \times 64$ and $15 \times 15 \times 64$ poses, respectively.
This results in $K_{train}=784$ and $K_{test}=28224$ on VIGOR, and $K_{train}=400$ and $K_{test}=14400$ on KITTI.

\subsection{Baselines}\label{subsec:baselines}

We compare SliceMatch to state-of-the-art global descriptor-based methods Cross-View Regression (CVR) \cite{vigor} and Multi-Class Classification (MCC)~\cite{zimin_metricloc} on the VIGOR dataset\footnote{We re-trained and evaluated the existing baselines on our corrected ground truth locations (see Section~\ref{subsec:datasets} and our Supplementary Material). The improved ground truth and code for our model are available at \url{https://github.com/tudelft-iv/SliceMatch}.}. 
Since CVR does localization with known orientation and, in~\cite{zimin_metricloc}, MCC mainly focuses on localization, we compare SliceMatch to baselines for localization with known orientation and also for 3-DoF pose estimation.
Following~\cite{zimin_metricloc}, we train CVR~\cite{vigor} for localization only (not retrieval) as it gives better localization results.
On the KITTI dataset, SliceMatch is compared to dense local feature-based fine-grained image retrieval method DSM~\cite{shi2020looking}, and to iterative camera pose estimation method LM~\cite{shi2022beyond}.
In~\cite{shi2022beyond}, the LM method is trained and tested with a 20\textdegree \ prior on the ground camera's orientation. 
We adopt the same setting and additionally provide the results with LM and SliceMatch trained and tested with unknown orientation.
On both datasets, baselines are trained with inputs with the same size as used for SliceMatch.

\subsection{Ablation Study}\label{subsec:ablation}

Before other experiments, we test on the VIGOR tuning set using $\alpha \in \{2,4,8,16,K\}$ for the loss of Equation~\eqref{eq:methodology_infonce_loss}, and tune the number of slices $N$.
We find $\alpha=4$ gives the best result, yielding 0.48m improvement on the mean localization error for our model compared to $\alpha=K$ in the original infoNCE loss~\cite{oord2018representation}.
If the number of slices $N$ is small,
the mean and median localization and orientation estimation errors increase (see Table~\ref{tab:number_slices}).
The model with $N=1$ cannot infer orientation.
When the width of the ground feature map $W$ is not a multiple of $N$, we interpolate the ground feature map $z_g$ to acquire $\sliceFeatG^n$.
However, it can be seen that the performance saturates above 16 slices.
Next, we tested SliceMatch without cross-view attention by dropping the concatenated $S^n$ in Equation~\eqref{eq:crossview_attention}.
Table~\ref{tab:number_slices} shows that including our proposed cross-view attention module brings a boost to both localization and orientation estimation performance.
Thus, we include cross-view attention and use $\alpha = 4$ and $N = 16$ in our main experiments.

\begin{table}[ht]
    \small
    \centering
    \begin{tabularx}{0.98\columnwidth}{L{0.8cm} R{1cm} R{1cm} R{1cm} R{1cm} R{1cm}}
        \toprule
        & \multicolumn{1}{|c|}{Cross-View} & \multicolumn{2}{c|}{↓ Location (m)} & \multicolumn{2}{c}{↓ Orientation (\textdegree)} \\[-0.2mm]
        N & \multicolumn{1}{|c|}{Attention} & \heading{Mean} & \multicolumn{1}{c|}{Median} & \heading{Mean} & \heading{Median} \\
        \midrule
        1 & \multicolumn{1}{|c|}{\sffamily X} & \multicolumn{1}{r}{12.73} & \multicolumn{1}{r|}{11.51} & \multicolumn{1}{r}{-\ \ \ \ } & \multicolumn{1}{r}{-\ \ \ \ } \\
        \midrule
        4 & \multicolumn{1}{|c|}{\checkmark} & \multicolumn{1}{r}{9.47} & \multicolumn{1}{r|}{7.47} & \multicolumn{1}{r}{51.49} & \multicolumn{1}{r}{32.96} \\
        8 & \multicolumn{1}{|c|}{\checkmark} & \multicolumn{1}{r}{9.16} & \multicolumn{1}{r|}{6.81} & \multicolumn{1}{r}{37.68} & \multicolumn{1}{r}{15.58} \\
        16 & \multicolumn{1}{|c|}{\checkmark} & \multicolumn{1}{r}{\textbf{7.60}} & \multicolumn{1}{r|}{\textbf{5.23}} & \multicolumn{1}{r}{\textbf{29.27}} & \multicolumn{1}{r}{\textbf{9.22}} \\
        32 & \multicolumn{1}{|c|}{\checkmark} & \multicolumn{1}{r}{8.14} & \multicolumn{1}{r|}{5.31} & \multicolumn{1}{r}{32.01} & \multicolumn{1}{r}{10.31} \\
        \midrule
        16$^\ddagger$ & \multicolumn{1}{|c|}{\checkmark} & \multicolumn{1}{r}{8.08} & \multicolumn{1}{r|}{5.44} & \multicolumn{1}{r}{31.05} & \multicolumn{1}{r}{11.02} \\
        \midrule
        16 & \multicolumn{1}{|c|}{\sffamily X} & \multicolumn{1}{r}{7.93} & \multicolumn{1}{r|}{5.81} & \multicolumn{1}{r}{29.50} & \multicolumn{1}{r}{12.32} \\
        \bottomrule
    \end{tabularx}
    \caption{\textbf{Location and orientation error for different slice number $N$ values on the VIGOR tuning split.}  $\ddagger$ indicates model trained with original infoNCE loss~\cite{oord2018representation}. Best performance in \textbf{bold}.}
    \label{tab:number_slices}
\end{table}

\subsection{Same-Area Generalization}\label{subsec:same-area}

\begin{table*}[htpb]
    \small
    \begin{tabularx}{0.9825\textwidth}{L{2.3cm} R{1cm} R{1cm} R{1cm} R{1cm} R{1cm} R{1cm} R{1cm} R{1cm} R{1cm} R{1cm}}
        \toprule
        & \multicolumn{1}{|c}{} & \multicolumn{1}{|c|}{} & \multicolumn{4}{c|}{Same-Area} & \multicolumn{4}{c}{Cross-Area} \\[0.5mm]
        
        & \multicolumn{1}{|c}{} & \multicolumn{1}{|c|}{Aligned} & \multicolumn{2}{c|}{↓ Location (m)} & \multicolumn{2}{c|}{↓ Orientation (\textdegree)} & \multicolumn{2}{c|}{↓ Location (m)} & \multicolumn{2}{c}{↓ Orientation (\textdegree)} \\[-0.2mm]
        
        Model & \multicolumn{1}{|c}{Backbone} & \multicolumn{1}{|c|}{Images} & \heading{Mean} & \multicolumn{1}{c|}{Median} & \heading{Mean} & \multicolumn{1}{c|}{Median} & \heading{Mean} & \multicolumn{1}{c|}{Median} & \heading{Mean} & \heading{Median} \\
        \midrule
        CVR~\cite{vigor} & \multicolumn{1}{|c}{VGG16} & \multicolumn{1}{|c|}{\checkmark} & 8.99 & \multicolumn{1}{r|}{7.81} & \multicolumn{1}{r}{-\ \ \ \ } & \multicolumn{1}{r|}{-\ \ \ \ } & 8.89 & \multicolumn{1}{r|}{7.73} & \multicolumn{1}{r}{-\ \ \ \ } & \multicolumn{1}{r}{-\ \ \ \ } \\
        MCC~\cite{zimin_metricloc} & \multicolumn{1}{|c}{VGG16} & \multicolumn{1}{|c|}{\checkmark} & 6.94 & \multicolumn{1}{r|}{3.64} & \multicolumn{1}{r}{-\ \ \ \ } & \multicolumn{1}{r|}{-\ \ \ \ } & 9.05 & \multicolumn{1}{r|}{5.14} & \multicolumn{1}{r}{-\ \ \ \ } & \multicolumn{1}{r}{-\ \ \ \ } \\
        SliceMatch (ours) & \multicolumn{1}{|c}{VGG16} & \multicolumn{1}{|c|}{\checkmark} & \textbf{5.18} & \multicolumn{1}{r|}{\textbf{2.58}} & \multicolumn{1}{r}{-\ \ \ \ } & \multicolumn{1}{r|}{-\ \ \ \ } & \textbf{5.53} & \multicolumn{1}{r|}{\textbf{2.55}} & \multicolumn{1}{r}{-\ \ \ \ } & \multicolumn{1}{r}{-\ \ \ \ } \\
        \midrule
        MCC~\cite{zimin_metricloc} & \multicolumn{1}{|c}{VGG16} & \multicolumn{1}{|c|}{{\sffamily X}} & 9.87 & \multicolumn{1}{r|}{6.25} & 56.86 & \multicolumn{1}{r|}{16.02} & 12.66 & \multicolumn{1}{r|}{9.55} & 72.13 & 29.97 \\
        SliceMatch (ours) & \multicolumn{1}{|c}{VGG16} & \multicolumn{1}{|c|}{{\sffamily X}} & \textbf{8.41} & \multicolumn{1}{r|}{\textbf{5.07}} & \textbf{28.43} & \multicolumn{1}{r|}{\textbf{5.15}} & \textbf{8.48} & \multicolumn{1}{r|}{\textbf{5.64}} & \textbf{26.20} & \multicolumn{1}{r}{\textbf{5.18}} \\
        \midrule
        SliceMatch (ours) & \multicolumn{1}{|c}{ResNet50} & \multicolumn{1}{|c|}{{\sffamily X}} & \textbf{6.49} & \multicolumn{1}{r|}{\textbf{3.13}} & \textbf{25.46} & \multicolumn{1}{r|}{\textbf{4.71}} & \textbf{7.22} & \multicolumn{1}{r|}{\textbf{3.31}} & \multicolumn{1}{r}{\textbf{25.97}} & \multicolumn{1}{r}{\textbf{4.51}} \\
        \bottomrule        
    \end{tabularx}
    \centering
    \caption{\textbf{Location and orientation estimation errors on VIGOR~\cite{vigor}.}
    \emph{Aligned Images} means the ground image orientation is known. For \emph{unaligned images}, the models estimate the 3-DoF ground camera pose.
    Best performance in \textbf{bold}.}
    \label{tab:vigor_results}
\end{table*}

\begin{table*}[ht]
    \small
    \centering
    \begin{tabularx}{\textwidth}{L{2.5cm} R{0.8cm} R{0.8cm} R{0.75cm} R{0.75cm} R{0.8cm} R{0.8cm} R{0.8cm} R{0.8cm} R{1cm} R{1cm} R{1cm} R{1cm}}
        \toprule
        & \multicolumn{1}{|c}{} & \multicolumn{1}{|c|}{} & \multicolumn{2}{c|}{↓ Location (m)} & \multicolumn{2}{c|}{↑ Lateral (\%)} & \multicolumn{2}{c|}{↑ Long. (\%)} & \multicolumn{2}{c|}{↓ Orien. (\textdegree)} & \multicolumn{2}{c}{↑ Orien. (\%)}\\[-0.2mm]
        Model & \multicolumn{1}{|c}{Area} & \multicolumn{1}{|c|}{Prior} & \heading{Mean} & \multicolumn{1}{c|}{Median} & \heading{r@1m} & \multicolumn{1}{c|}{r@5m} & \heading{r@1m} & \multicolumn{1}{c|}{r@5m} & \multicolumn{1}{c}{Mean} & \multicolumn{1}{c|}{Median} & \heading{r@1\textdegree} & \heading{r@5\textdegree} \\
        \midrule
        DSM~\cite{shi2020looking} & \multicolumn{1}{|c}{Same} & \multicolumn{1}{|c|}{20\textdegree} & \multicolumn{1}{r}{-\ \ \ \ } & \multicolumn{1}{r|}{-\ \ \ \ } & \multicolumn{1}{r}{10.12} & \multicolumn{1}{r|}{48.24} & \multicolumn{1}{r}{4.08} & \multicolumn{1}{r|}{20.14} & \multicolumn{1}{r}{-\ \ \ \ } & \multicolumn{1}{r|}{-\ \ \ \ } & \multicolumn{1}{r}{3.58} & \multicolumn{1}{r}{24.44} \\
        LM~\cite{shi2022beyond} & \multicolumn{1}{|c}{Same} & \multicolumn{1}{|c|}{20\textdegree} & \multicolumn{1}{r}{12.08} & \multicolumn{1}{r|}{11.42} & \multicolumn{1}{r}{35.54} & \multicolumn{1}{r|}{80.36} & \multicolumn{1}{r}{5.22} & \multicolumn{1}{r|}{26.13} & \multicolumn{1}{r}{\textbf{3.72}} & \multicolumn{1}{r|}{\textbf{2.83}} & \multicolumn{1}{r}{\textbf{19.64}} & \multicolumn{1}{r}{\textbf{71.72}} \\
        SliceMatch (ours) & \multicolumn{1}{|c}{Same} & \multicolumn{1}{|c|}{20\textdegree} & \textbf{7.96} & \multicolumn{1}{r|}{\textbf{4.39}} & \multicolumn{1}{r}{\textbf{49.09}} & \multicolumn{1}{r|}{\textbf{98.52}} & \textbf{15.19} & \multicolumn{1}{r|}{\textbf{57.35}} & \multicolumn{1}{r}{4.12} & \multicolumn{1}{r|}{3.65} & \multicolumn{1}{r}{13.41} & \multicolumn{1}{r}{64.17} \\
        \midrule
        LM~\cite{shi2022beyond} & \multicolumn{1}{|c}{Same} & \multicolumn{1}{|c|}{\sffamily X} & \multicolumn{1}{r}{15.51} & \multicolumn{1}{r|}{15.97} & \multicolumn{1}{r}{5.17} & \multicolumn{1}{r|}{25.44} & \multicolumn{1}{r}{4.66} & \multicolumn{1}{r|}{25.39} & \multicolumn{1}{r}{89.91} & \multicolumn{1}{r|}{90.75} & \multicolumn{1}{r}{0.61} & \multicolumn{1}{r}{2.89} \\
        SliceMatch (ours) & \multicolumn{1}{|c}{Same} & \multicolumn{1}{|c|}{{\sffamily X}} & \textbf{9.39} & \multicolumn{1}{r|}{\textbf{5.41}} & \multicolumn{1}{r}{\textbf{39.73}} & \multicolumn{1}{r|}{\textbf{87.92}} & \textbf{13.63} & \multicolumn{1}{r|}{\textbf{49.22}} & \multicolumn{1}{r}{\textbf{8.71}} & \multicolumn{1}{r|}{\textbf{4.42}} & \multicolumn{1}{r}{\textbf{11.35}} & \multicolumn{1}{r}{\textbf{55.82}} \\
        \midrule
        DSM~\cite{shi2020looking} & \multicolumn{1}{|c}{Cross} & \multicolumn{1}{|c|}{20\textdegree} & \multicolumn{1}{r}{-\ \ \ \ } & \multicolumn{1}{r|}{-\ \ \ \ } & \multicolumn{1}{r}{10.77} & \multicolumn{1}{r|}{48.24} & \multicolumn{1}{r}{3.87} & \multicolumn{1}{r|}{19.50} & \multicolumn{1}{r}{-\ \ \ \ } & \multicolumn{1}{r|}{-\ \ \ \ } & \multicolumn{1}{r}{3.53} & \multicolumn{1}{r}{23.95} \\
        LM~\cite{shi2022beyond} & \multicolumn{1}{|c}{Cross} & \multicolumn{1}{|c|}{20\textdegree} & \multicolumn{1}{r}{\textbf{12.58}} & \multicolumn{1}{r|}{12.11} & \multicolumn{1}{r}{27.82} & \multicolumn{1}{r|}{72.89} & \multicolumn{1}{r}{5.75} & \multicolumn{1}{r|}{26.48} & \multicolumn{1}{r}{\textbf{3.95}} & \multicolumn{1}{r|}{\textbf{3.03}} & \multicolumn{1}{r}{18.42} & \multicolumn{1}{r}{\textbf{71.00}} \\
        SliceMatch (ours) & \multicolumn{1}{|c}{Cross} & \multicolumn{1}{|c|}{20\textdegree} & 13.50 & \multicolumn{1}{r|}{\textbf{9.77}} & \multicolumn{1}{r}{\textbf{32.43}} & \multicolumn{1}{r|}{\textbf{86.44}} & \textbf{8.30} & \multicolumn{1}{r|}{\textbf{35.57}} & \multicolumn{1}{r}{4.20} & \multicolumn{1}{r|}{6.61} & \multicolumn{1}{r}{\textbf{46.82}} & \multicolumn{1}{r}{46.82} \\
        \midrule
        LM~\cite{shi2022beyond} & \multicolumn{1}{|c}{Cross} & \multicolumn{1}{|c|}{\sffamily X} & \multicolumn{1}{r}{15.50} & \multicolumn{1}{r|}{16.02} & \multicolumn{1}{r}{5.60} & \multicolumn{1}{r|}{25.60} & \multicolumn{1}{r}{5.64} & \multicolumn{1}{r|}{25.76} & \multicolumn{1}{r}{89.84} & \multicolumn{1}{r|}{89.85} & \multicolumn{1}{r}{0.60} & \multicolumn{1}{r}{2.65} \\
        SliceMatch (ours) & \multicolumn{1}{|c}{Cross} & \multicolumn{1}{|c|}{{\sffamily X}} & \textbf{14.85} & \multicolumn{1}{r|}{\textbf{11.85}} & \multicolumn{1}{r}{\textbf{24.00}} & \multicolumn{1}{r|}{\textbf{72.89}} & \textbf{7.17} & \multicolumn{1}{r|}{\textbf{33.12}} & \multicolumn{1}{r}{\textbf{23.64}} & \multicolumn{1}{r|}{\textbf{7.96}} & \multicolumn{1}{r}{\textbf{31.69}} & \multicolumn{1}{r}{\textbf{31.69}} \\
        \bottomrule
    \end{tabularx}
    \caption{\textbf{Location and orientation estimation error and recall on KITTI~\cite{geiger2013vision, shi2022beyond}.} \emph{Prior} means the orientation is known with a certain amount of noise.
    \emph{Long.} and \emph{Orien.} are abbreviations for \emph{Longitudinal} and \emph{Orientation}, respectively. Best performance in \textbf{bold}. The results for DSM~\cite{shi2020looking} are taken from~\cite{shi2022beyond} and we used the trained LM model provided by~\cite{shi2022beyond} for its evaluation.
    }
    \label{tab:kitti_results}
\end{table*}

We test model generalization to new panoramic and limited HFoV ground images within the same area on VIGOR and KITTI.
As shown in Table~\ref{tab:vigor_results} Same-Area, SliceMatch surpasses CVR~\cite{vigor} and MCC~\cite{zimin_metricloc} in terms of both localization with known orientation and 3-DoF camera pose estimation.
Compared to MCC, in which location-wise discriminative features are learned, SliceMatch contrasts the learned global descriptors with aerial descriptors at different locations and orientations.
Hence, it is more discriminative especially w.r.t.~orientations, and has a 19\% and 68\% reduction in the median localization and median orientation error when the orientation of test ground images is unknown.
We use VGG16 as our main backbone for a fair comparison to the baselines,
though we note our localization and orientation error decreases even further when using ResNet50 as backbone.
% 19% = (6.25-5.07)/6.25, VIGOR same area pose estimation median localization error, SliceMatch vs. MCC
% 68% = (16.02-5.15)/16.02, VIGOR same area pose estimation median orientation error, SliceMatch vs. MCC
We show in Figure~\ref{fig:slicematch_visuals}b that SliceMatch can express its multimodal uncertainty when the observed scene has a symmetric layout.
However, it sometimes picks candidate poses at a wrong mode, resulting in large errors (see Figure~\ref{fig:slicematch_visuals}c).
Over all test samples, SliceMatch has a substantially lower median error than its mean error for both localization and orientation estimation, indicating that the mean is skewed by such outliers.
In practice, SliceMatch's multimodal uncertainty could be resolved by applying downstream a probabilistic temporal filter on its output~\cite{zimin_metricloc}.

As shown in Table~\ref{tab:kitti_results} Same-Area, on the KITTI dataset, both camera pose estimation methods, LM~\cite{shi2022beyond} and SliceMatch, surpass the fine-grained image retrieval-based method DSM~\cite{shi2020looking}.
When the orientation prior is present, SliceMatch has 34\% and 62\% lower mean and median localization error than LM~\cite{shi2022beyond}, and its recall@1m and recall@5m is higher than that of LM~\cite{shi2022beyond} for localization in both lateral and longitudinal directions.
% 34% = (12.08-7.96)/12.08, KITTI same area pose estimation mean localization error with 20 degrees prior, SliceMatch vs. LM
% 62% = (11.42-4.39)/11.42, KITTI same area pose estimation median localization error with 20 degrees prior, SliceMatch vs. LM
Notably, since the ground images in KITTI view in the driving direction with a limited HFoV, finding the location along the longitudinal direction is more challenging than that for the lateral direction.
Thus, recall for longitudinal direction is considerably lower than that for lateral direction, and this trend applies to all compared methods.
The iterative refinement LM method~\cite{shi2022beyond} shows its advantage in orientation prediction when the strong orientation prior is present.
We highlight that SliceMatch can work without this prior.
In contrast, LM~\cite{shi2022beyond} relies on the projection of dense local features from the aerial view to ground view~\cite{shi2022beyond} and does not work when there is no same scene captured in the projected view and the ground view (see Table~\ref{tab:kitti_results}).

\subsection{Cross-Area Generalization}
Generalization to new ground images in different areas is a more difficult task than that in the same area since the test area can look very different from the training area (e.g. different cities in the VIGOR dataset).
As shown in Table~\ref{tab:vigor_results} Cross-Area, SliceMatch generalizes well under this challenging setting in terms of both localization and orientation estimation, while we observe more degeneration in the cross-area test performance of MCC~\cite{zimin_metricloc}.
MCC's feature decoder receives the full scene information from its encoder, while SliceMatch divides the observed scene into slices and seeks per-slice discriminative features, resulting in more robustness against the change of the scene.
Again, using a ResNet50 backbone further improves our results.

On KITTI Test2 set (Table~\ref{tab:kitti_results} Cross-Area), SliceMatch achieves a lower median localization error than LM~\cite{shi2022beyond} when the 20\textdegree \ orientation prior is present in both training and testing.
But our mean error is higher than LM~\cite{shi2022beyond} by 0.92m
and LM~\cite{shi2022beyond} surpasses SliceMatch in orientation prediction when a strong prior is available.
SliceMatch performs considerably better when no orientation prior is available
as LM~\cite{shi2022beyond} gets stuck in local optima.

\subsection{Runtime Analysis}\label{subsec:runtime_analysis}
We compare the runtime of SliceMatch to that of baselines on the same hardware, a single NVIDIA Tesla V100 GPU.
For all baselines, we use the released code from their authors.
CVR~\cite{vigor} and MCC~\cite{zimin_metricloc} are implemented in TensorFlow, LM~\cite{shi2022beyond} and our SliceMatch in PyTorch.
The frames per second (FPS) are calculated by taking the average inference time per input pair over all test samples.
On VIGOR, SliceMatch achieves an FPS of 167, which is considerably faster than global descriptor-based baselines: 50~FPS for CVR~\cite{vigor} for localization only, 29~FPS / 3~FPS for MCC~\cite{zimin_metricloc} for localization only / pose estimation.
On KITTI, SliceMatch runs at 156 FPS, while the local feature-based iterative method, LM~\cite{shi2022beyond} has 0.59~FPS.
% \commentOLD{To infer on a pair of input images on the KITTI dataset, SliceMatch achieved 156~FPS 6.4ms (156~FPS), while local feature-based iterative method, LM~\cite{shi2022beyond}, takes 1.7s (0.59~FPS).}
Importantly, the runtime of SliceMatch remains nearly constant as the number of used candidate poses $K$ increases (we tested $K$ up to \num{1e6}), see details in Supplementary Material.
% \commentOLD{In practice, memory will thus be the limiting factor for determining the number of poses.}

%% file: 5-conclusion.tex
\section{Conclusion}\label{sec:conclusion}

We have introduced SliceMatch, a novel, accurate, and efficient method for cross-view 3-DoF camera pose estimation.
By splitting the HFoV into slices, our architecture can learn discriminative features in terms of both localization and orientation estimation. 
Our proposed aggregation
can select the relevant aerial image features for each ground view slice through cross-view attention,
and we observe further accuracy gains by reweighing the terms in the infoNCE loss.
With the same VGG backbone,
SliceMatch achieves 19\% and 62\% lower median localization error than the previous state-of-the-art on the VIGOR and KITTI datasets.
A better backbone improves SliceMatch's performance even further, e.g.~with ResNet50 its 50\% lower median error on VIGOR sets a new state-of-the-art.
% 19% = (6.25-5.07)/6.25, VIGOR same area pose estimation SliceMatch vs. MCC
% 62% = (11.42-4.39)/11.42, KITTI same area pose estimation with 20 degrees prior, SliceMatch vs. LM
To construct the global descriptor for a candidate pose,
only an efficient weighted averaging over the aerial features is needed using precomputed masks (which represent the ground camera's frustum geometry in the aerial view),
achieving inference at more than 150 FPS.
SliceMatch can include available priors in its candidate poses, e.g.~for an initial orientation estimate, but does not require it.
Future work will explore adapting pose candidates for temporal filtering and sensor fusion.

%% file: supp-overview.tex
\section*{Overview}\label{sec:overview}

\noindent In this supplementary material, we provide the following items for a better understanding of the main paper: \\
% \commentZX{You can refer to the sections in the main paper (without hyperlinks).}
% \commentZX{I would suggest using sections A, B, C, etc. for the sections in the appendix to avoid confusion on the section index of the main paper.}

\noindent \ \ A. Ground Truth Labels of VIGOR Dataset (\emph{Section~\ref{subsec:datasets}}) % 4.1

\noindent \ \ B. Tuning Number of Slices (\emph{Section~\ref{subsec:ablation}}) % 4.5

\noindent \ \ C. Inference on Images with a Limited HFoV (\emph{Section~\ref{subsec:same-area}}) % 4.6

\noindent \ \ D. Details on Runtime and Memory Usage (\emph{Section~\ref{subsec:runtime_analysis}}) % 4.8

\noindent \ \ E. Visualization: SliceMatch Predictions (\emph{Section~\ref{subsec:same-area}}) % 4.6

%% file: supp-text1.tex
\section*{A. Ground Truth Labels of VIGOR Dataset}\label{sec:vigor_labels}

\begin{figure}[t]
    \centering
    \subfloat[\centering Ground view]{{\includegraphics[height=2.5cm]{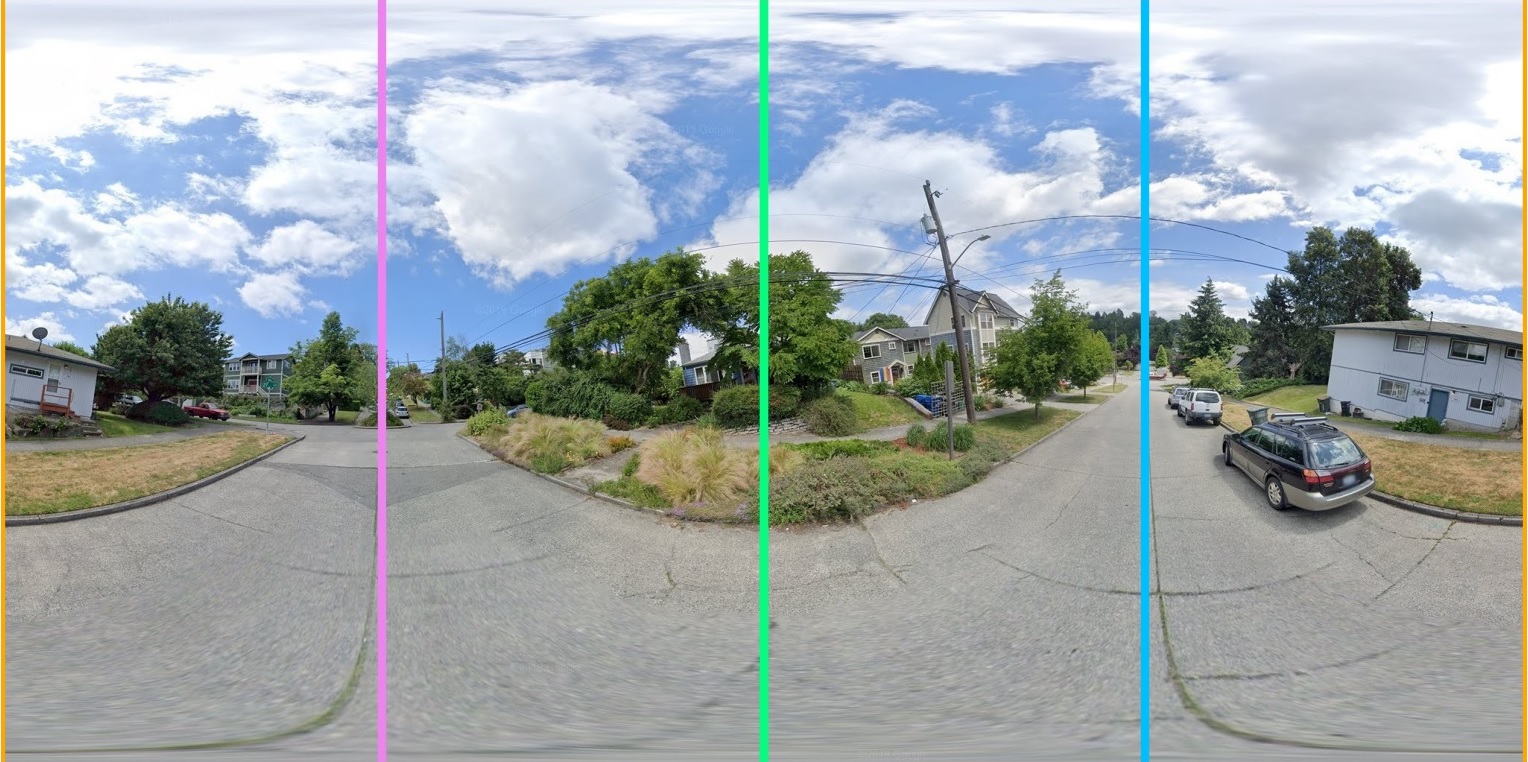}}}
    \hspace{-5mm}%
    \qquad
    \subfloat[\centering Positive]{{\includegraphics[height=2.5cm]{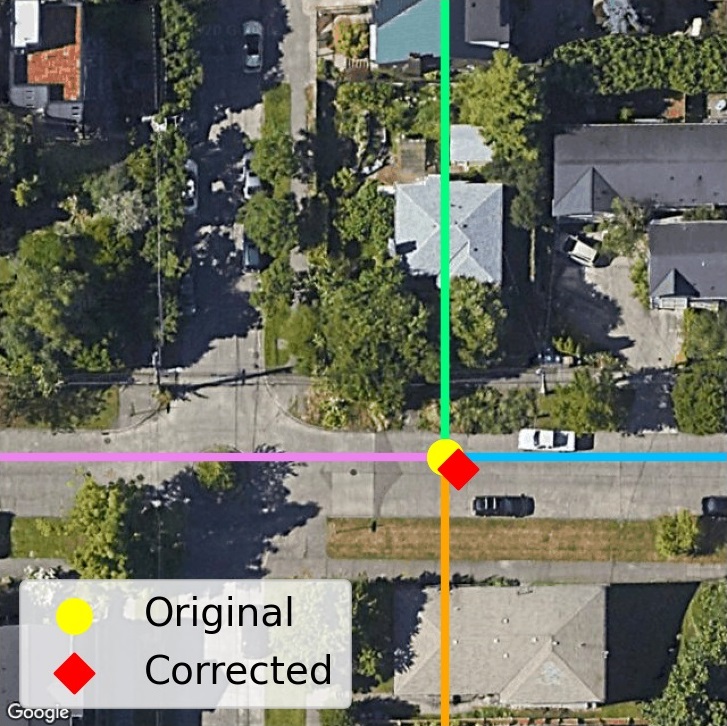}}}
    \qquad
    \subfloat[\centering Semi-pos. 1]{{\includegraphics[height=2.5cm]{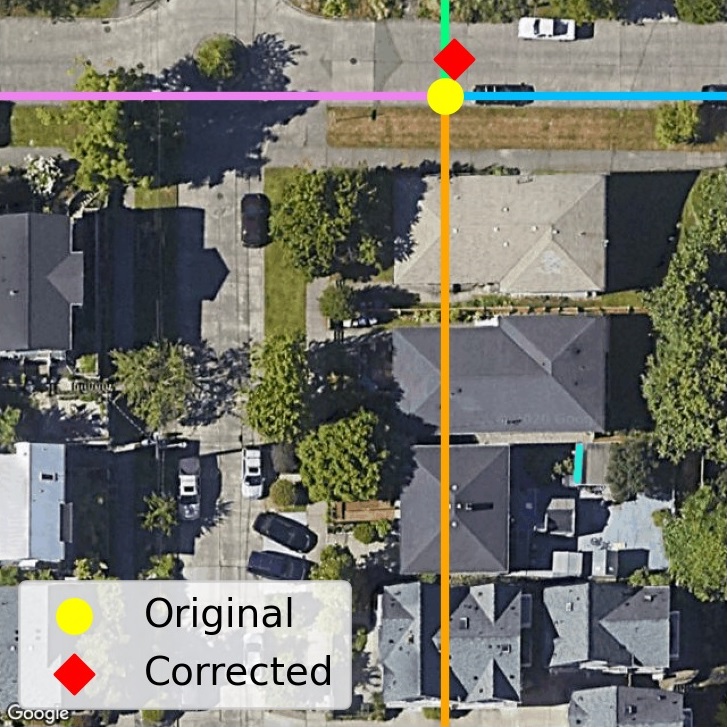}}}
    \hspace{-5mm}%
    \qquad
    \subfloat[\centering Semi-pos. 2]{{\includegraphics[height=2.5cm]{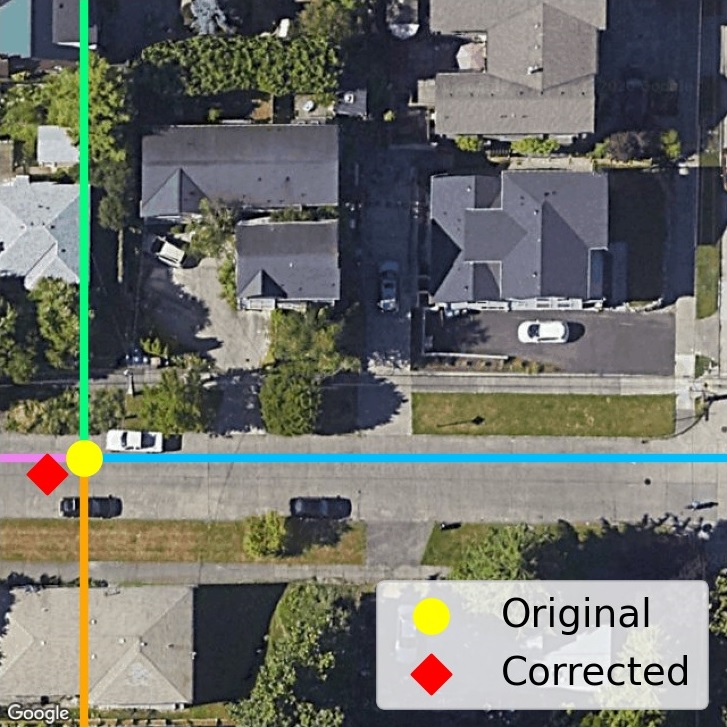}}}
    \hspace{-5mm}%
    \qquad
    \subfloat[\centering Semi-pos. 3]{{\includegraphics[height=2.5cm]{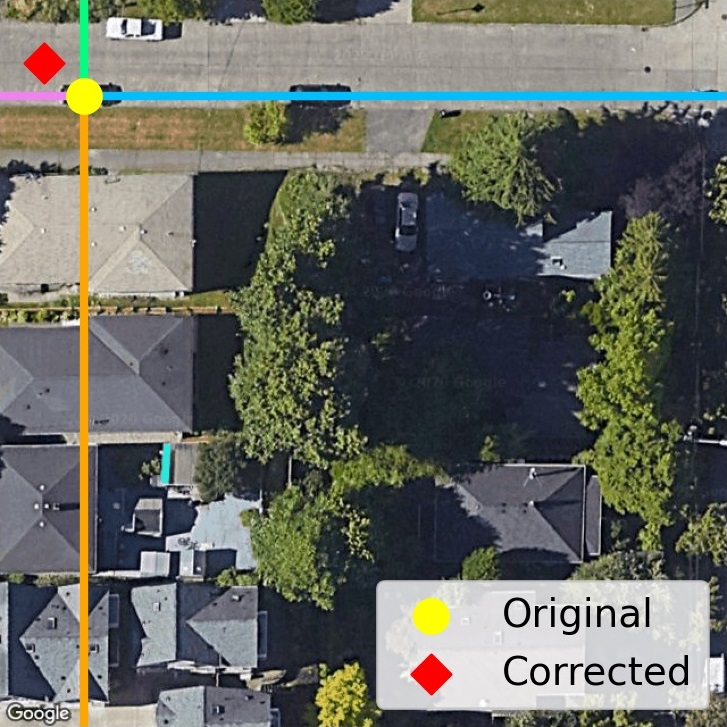}}}
    \caption{\textbf{A ground image together with the four matching aerial images from VIGOR~\cite{vigor}.} The original and the corrected locations are indicated by the yellow dot and red diamond, respectively. South, West, North, and East are the orange, pink, green, and blue lines, respectively. \emph{Semi-pos.} means \emph{Semi-positive}.}
    \label{fig:vigor_image_pair}
\end{figure}

We have visually inspected the image pairs of the VIGOR dataset~\cite{vigor} and noticed a location inconsistency between image pairs that share the ground image. Figure~\ref{fig:vigor_image_pair} shows an image pair formed by a ground image and a positive or semi-positive aerial image from Seattle with the original and corrected ground truth camera locations indicated. Depending on the aerial image, the original ground truth location (yellow dot) is in different locations. However, this should be the same visual location (red diamond) for all aerial images corresponding to this specific ground image.

\begin{figure*}[t]
    \centering
    \subfloat[\centering Aerial image 1]{{\includegraphics[height=4.5cm]{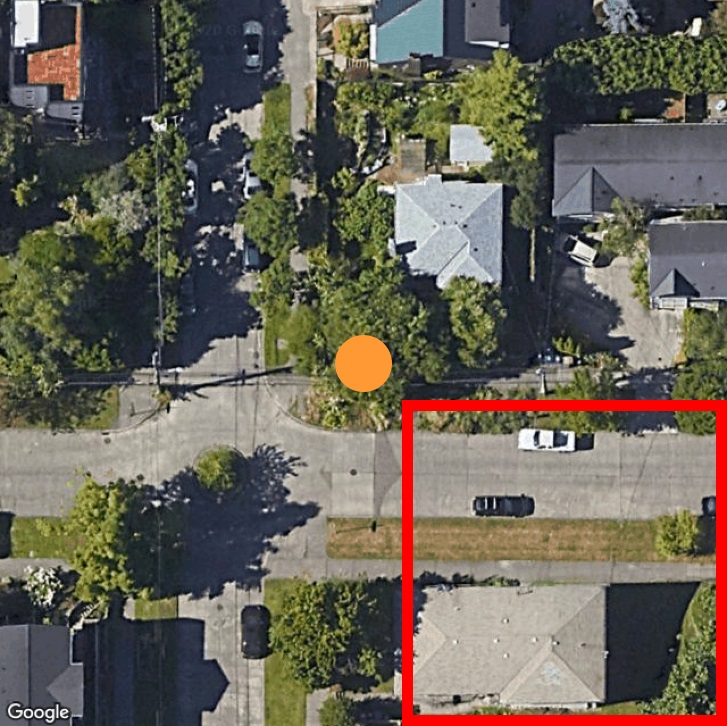}}}
    \hspace{5mm}%
    \qquad
    \subfloat[\centering Aerial image 2]{{\includegraphics[height=4.5cm]{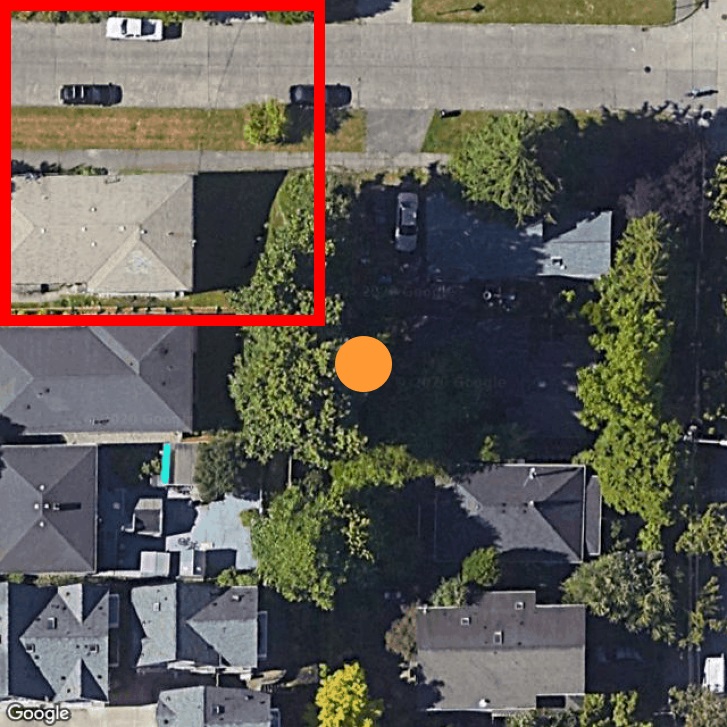}}}
    \hspace{5mm}%
    \qquad
    \subfloat[\centering The overlaid aerial image 1 and 2]{{\includegraphics[height=4.5cm]{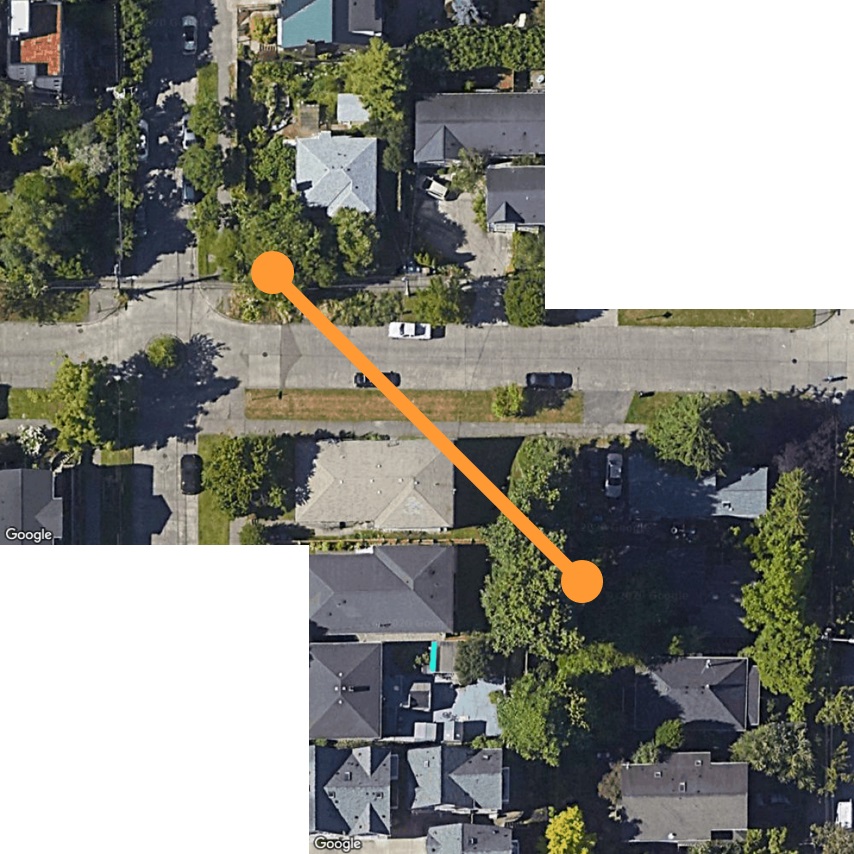}}}
    \caption{\textbf{Two aerial images from the VIGOR dataset~\cite{vigor} that have a visual overlap.} Each image center is indicated by an orange dot and the connection between the two dots shows the distance in pixels. (a) Aerial image 1 with the visual overlap indicated by the red contour. (b) Aerial image 2 with the visual overlap indicated by the red contour. (c) The overlaid aerial image 1 and 2. We use cross-correlation to find the amount of overlapping (in pixels) between aerial image 1 and 2.}
    \label{fig:vigor_cross_correlation}
\end{figure*}

The authors of the VIGOR dataset~\cite{vigor} have used a ground resolution equal to 0.114m/pixel for all 4 cities of the dataset to convert the latitude and longitude of a ground image to its location in aerial images.
We have measured the ground resolution ourselves. 
Pixel-level correspondences between aerial images that have a visual overlap can be calculated using cross-correlation.
Then we can overlay these aerial images.
Figure~\ref{fig:vigor_cross_correlation} shows this for two aerial images. The distance in pixels between the two image centers can be measured (see Figure~\ref{fig:vigor_cross_correlation}c). In addition, the longitude and latitude of the image center of each aerial image are known, allowing the distance to be determined in meters as well. The ground resolution of an aerial-aerial image combination can be calculated using,

\begin{equation}
\begin{aligned}
\textit{ground} \ \textit{resolution} = \frac{\textit{distance} \ \textit{in} \ \textit{meters}}{\textit{distance} \ \textit{in} \ \textit{pixels}}.
\end{aligned}
\label{eq:ground_resolution}
\end{equation}

% Wrong ground truth labels
% Cross-correlation

% Cross-correlation

We have calculated a new ground resolution for each city by averaging the ground resolutions of a city's aerial-aerial image combinations. Table~\ref{tab:vigor_ground_resolutions} shows the original and our measured ground resolutions. It turns out that the ground resolutions differ per city. The measured value for the ground resolution for New York is almost equal to the ground resolution that the VIGOR authors have used. However, the measured resolutions for the other cities differ significantly.

The measured ground resolutions have been used to determine corrected ground truth location labels. Table~\ref{tab:appendix_statistics} shows statistics on the absolute error in meters between the original and corrected locations. The positive image pairs of the dataset were used to determine the statistics since only those image pairs were used for the experiments (see Section~\ref{subsec:datasets}).
For Seattle, the difference between the original and measured ground resolution is the largest and this results in errors of more than 3 meters (see Table~\ref{tab:vigor_ground_resolutions}). The other 3 cities have smaller mean and median errors than Seattle.

In our localization only and pose estimation experiments for the VIGOR dataset, we resize the aerial image to $512\times512$ pixels. As a result, the ground resolution of the resized aerial images can be obtained by multiplying the measured ground resolutions from Table~\ref{tab:vigor_ground_resolutions} by 1.25 ($=640/512$).

\vspace{1mm}

% Ground resolutions
\begin{table}[ht]
    \centering
    \begin{tabularx}{0.85\columnwidth}{L{3cm} R{1cm} R{1cm}}
        \toprule
        City & \multicolumn{1}{|c|}{Original} & \heading{Measured} \\
        \midrule
        Chicago & \multicolumn{1}{|r|}{0.114} & \multicolumn{1}{r}{0.111} \\
        New York & \multicolumn{1}{|r|}{0.114} & \multicolumn{1}{r}{0.113} \\
        San Francisco & \multicolumn{1}{|r|}{0.114} & \multicolumn{1}{r}{0.118} \\
        Seattle & \multicolumn{1}{|r|}{0.114} & \multicolumn{1}{r}{0.101} \\
        \bottomrule
    \end{tabularx}
    \caption{\textbf{The original and our measured ground resolution for the 4 cities from VIGOR~\cite{vigor}.} The ground resolutions correspond to aerial images with a size of $640\times640$ pixels, and the unit of the ground resolution is m/pixel.}
    \label{tab:vigor_ground_resolutions}
\end{table}

\vspace{-2mm}

% Absolute errors (statistics)
\begin{table}[ht]
    \centering
    \begin{tabularx}{0.95\columnwidth}{L{2cm} R{1cm} R{1cm} R{1cm} R{1cm} R{1cm} R{1cm} R{1cm} R{1cm}}
        \toprule
        City & \multicolumn{1}{|c}{Min.} & \heading{Mean} & \heading{Median} & \heading{Max.} \\
        \midrule
        Chicago & \multicolumn{1}{|c}{0.00} & 0.43 & 0.45 & 0.80 \\
        New York & \multicolumn{1}{|c}{0.00} & 0.25 & 0.25 & 0.47 \\
        San Francisco & \multicolumn{1}{|c}{0.00} & 0.46 & 0.49 & 0.95 \\
        Seattle & \multicolumn{1}{|c}{0.00} & 1.72 & 1.79 & 3.14 \\
        \bottomrule
    \end{tabularx}
    \caption{\textbf{Statistics on the absolute error in meters of the labels for the 4 cities from VIGOR~\cite{vigor}.} The absolute error is defined as the distance between the original and the corrected locations. \emph{Min.} and \emph{Max.} indicate \emph{Minimum} and \emph{Maximum}, respectively.}
    \label{tab:appendix_statistics}
\end{table}

\vspace{-2mm}

%% file: supp-text2.tex
\section*{B. Tuning Number of Slices}\label{sec:KITTI_slice_number}

\begin{figure*}[th]%
    \centering%
    \includegraphics[width=4cm]{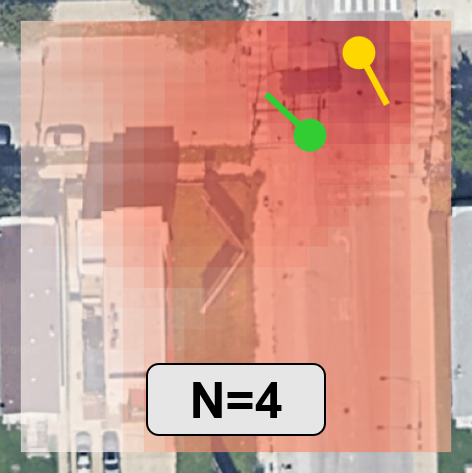} \quad
    \includegraphics[width=4cm]{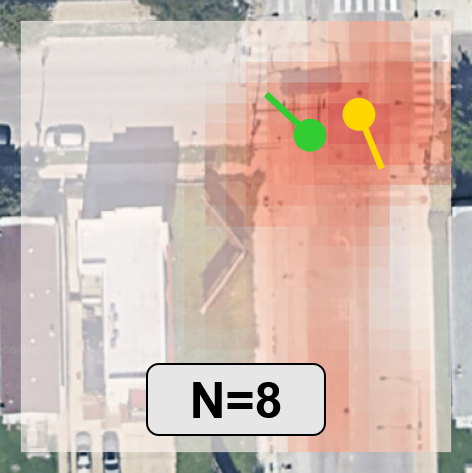} \quad
    \includegraphics[width=4cm]{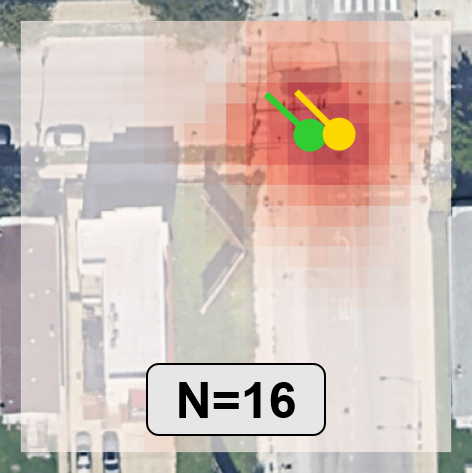}
    \caption{\textbf{SliceMatch models with different slice number $N$ values.} The ground truth camera pose and our estimated camera pose are in green and yellow, respectively. Red shading indicates the highest similarity score between the ground descriptor and the aerial descriptors among all orientations at that location.}
    \label{fig:heatmaps}
\end{figure*}

To supplement our ablation study on the number of slices  (see Section~\ref{subsec:ablation}), we visualize the predictions from SliceMatch models with different numbers of slices on VIGOR same-area, see Figure~\ref{fig:heatmaps}.
Using larger $N$ (more slices) maintains more of the relative orientation between the visible components in the scene.
Note that lowering $N$ makes the descriptors less orientation aware, which we observe lowers performance.
Generally, we observe that increasing to $N=16$ makes the descriptors more discriminative,
resulting in less uncertainty about the true location and orientation.
However, our ablation study in the main paper Table 1 demonstrates a trade-off: if $N$ becomes too large, the descriptor becomes too sensitive to pose differences between the best candidate pose and true pose.

Similarly, we also conducted an ablation study for the number of slices on the KITTI dataset~\cite{geiger2013vision, shi2022beyond}, see Table~\ref{tab:KITTI_number_slices}. 
For this study, we used the Same-Area setting of KITTI and the 20\textdegree \ orientation prior. 
Similar to the VIGOR dataset, the highest performance is achieved with 16 slices for the KITTI dataset as well. For 8 and 32 slices, the performance is slightly worse.

\begin{table}[h]
    \small
    \centering
    \begin{tabularx}{0.98\columnwidth}{L{0.8cm} R{1cm} R{1cm} R{1cm} R{1cm} R{1cm}}
        \toprule
        & \multicolumn{1}{|c|}{Cross-View} & \multicolumn{2}{c|}{↓ Location (m)} & \multicolumn{2}{c}{↓ Orientation (\textdegree)} \\[-0.2mm]
        N & \multicolumn{1}{|c|}{Attention} & \heading{Mean} & \multicolumn{1}{c|}{Median} & \heading{Mean} & \heading{Median} \\
        \midrule
        8 & \multicolumn{1}{|c|}{\checkmark} & \multicolumn{1}{r}{8.74} & \multicolumn{1}{r|}{5.11} & \multicolumn{1}{r}{4.54} & \multicolumn{1}{r}{4.01} \\
        16 & \multicolumn{1}{|c|}{\checkmark} & \multicolumn{1}{r}{\textbf{7.96}} & \multicolumn{1}{r|}{\textbf{4.39}} & \multicolumn{1}{r}{\textbf{4.12}} & \multicolumn{1}{r}{\textbf{3.65}} \\        
        32 & \multicolumn{1}{|c|}{\checkmark} & \multicolumn{1}{r}{8.03} & \multicolumn{1}{r|}{4.72} & \multicolumn{1}{r}{4.34} & \multicolumn{1}{r}{3.65} \\
        \bottomrule
    \end{tabularx}
    \caption{\textbf{Location and orientation error for different slice number $N$ values for the Same-Area setting on the KITTI dataset~\cite{geiger2013vision, shi2022beyond}.} Best performance in \textbf{bold}.}
    \label{tab:KITTI_number_slices}
\end{table}

%% file: supp-text3.tex
\section*{C. Inference on Images with a Limited HFoV}\label{sec:limited_HFoV}

\begin{figure}[h]
    \centering
    \subfloat[\centering Location estimation performance]{{\includegraphics[width=0.9\columnwidth]{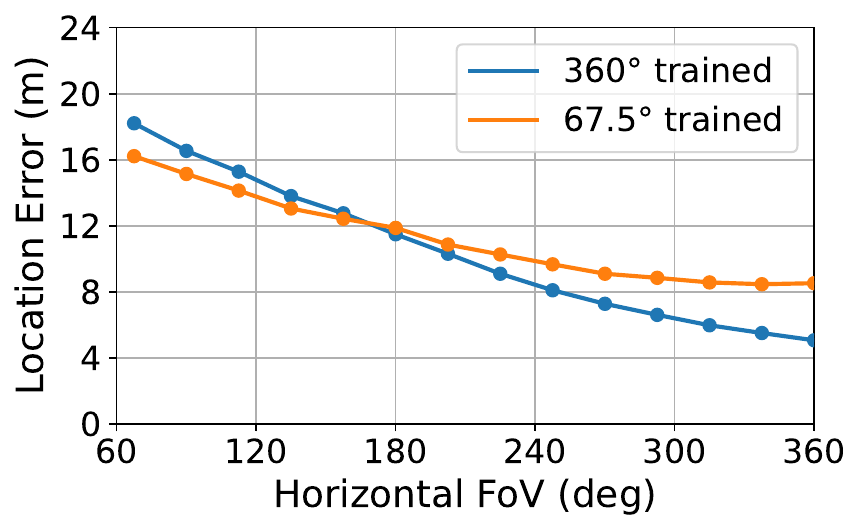}}}
    \qquad
    \subfloat[\centering Orientation estimation performance]{{\includegraphics[width=0.9\columnwidth]{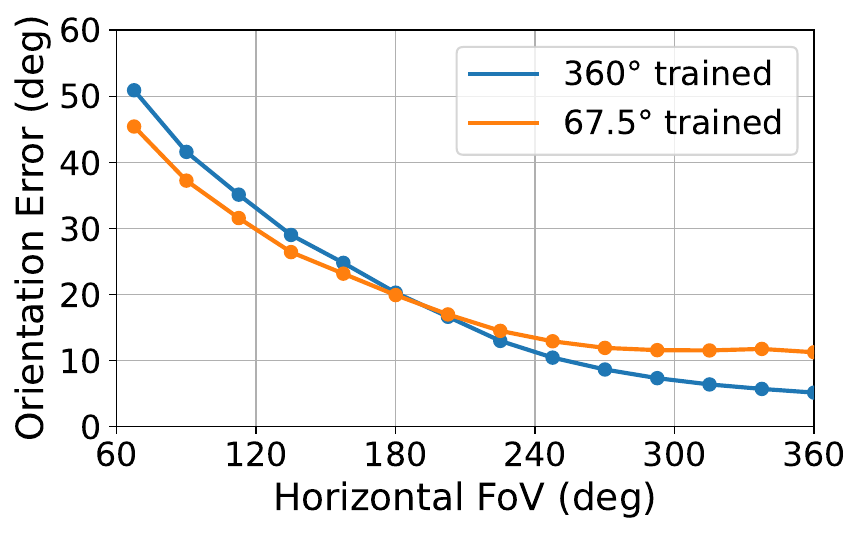}}}
    \caption{\textbf{Median location and orientation estimation errors on VIGOR~\cite{vigor} for limited HFoV.} (a) Location estimation performance. (b) Orientation estimation performance.}
    \label{fig:hfov}
\end{figure}

Additionally, we conducted experiments that vary the HFoV of test images in the VIGOR dataset (same-area), see Figure~\ref{fig:hfov} (median errors).
As expected, SliceMatch's performance degrades when the HFoV of the ground-level query image reduces, as it contains less information.
Training on ground images with a small HFoV, e.g. $\sim 67.5^\circ$, recuperates some performance when testing on small HFoVs.

%% file: supp-text4.tex
\section*{D. Details on Runtime and Memory Usage}\label{sec:runtime}

Here, we provide a detailed analysis of the runtime and memory usage of SliceMatch (see Section~\ref{subsec:runtime_analysis}). % 4.8

Both pre-computation of slice masks and parallelization of pose descriptor aggregation contribute to our efficiency. 
For a single input pair, we process candidate poses in parallel by performing a single (large) matrix multiplication (the matrix has the pre-computed masks as rows). Note that the candidate poses and thus the masks are the same for all test images. Since we never need to recompute the masks, pre-computation is excluded from the reported inference time. 
On a single input pair in the VIGOR dataset, our \textit{feature extraction} / \textit{descriptors aggregation} / \textit{pose comparison} takes 3.4ms / 1.5ms / 1.1ms, respectively. For the KITTI dataset, this is 3.6ms / 1.6ms / 1.2ms, respectively.

% For selecting candidate poses at each candidate location in the aerial image, we prefer to construct $M \cdot N$ azimuth directions at a fixed interval of $360^\circ/(M \cdot N)$, where $N$ is the number of slices in the ground view.
% In this case, the pre-constructed slice masks can be reused by candidate poses with different orientation angles, resulting in $M \cdot N$ unique slice masks at each candidate location other than $M \cdot N^2$.

\begin{figure}[h]
    \centering
    \subfloat[\centering Frames per second]{{\includegraphics[height=4cm]{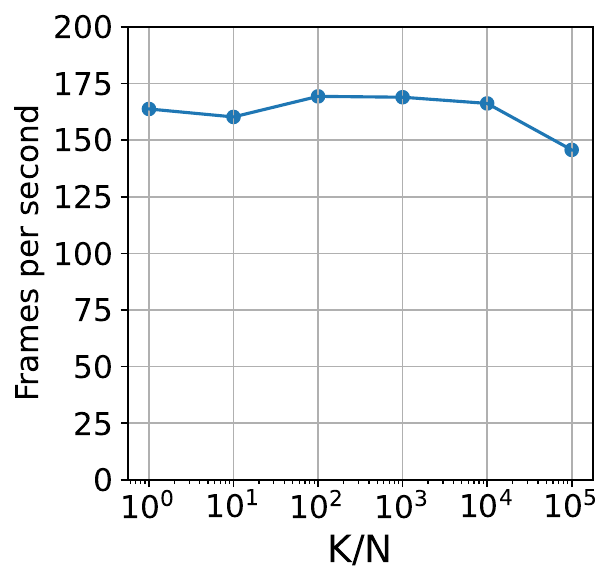}}}
    \subfloat[\centering Memory usage]{{\includegraphics[height=4cm]{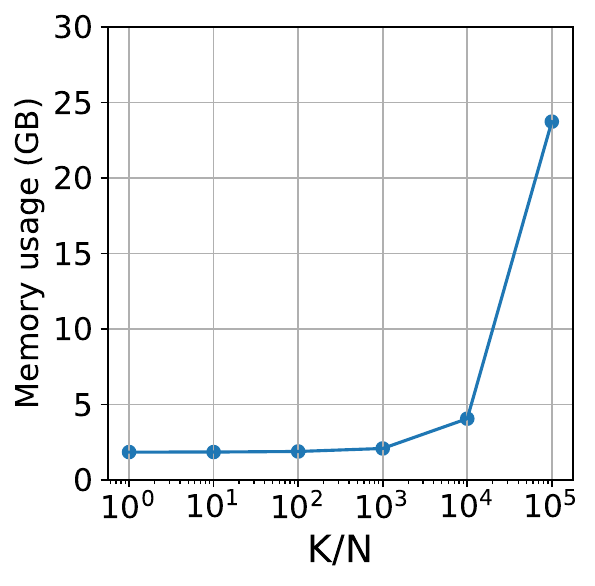}}}
    \caption{\textbf{SliceMatch frames per second and memory usage for a varying number of poses.} Note that the x-axis uses log-scale.}
    \label{fig:computational_efficiency}
\end{figure}

Importantly, the runtime of SliceMatch remains nearly constant as the number of used candidate poses $K$ increases, while memory scales linearly, see Figure~\ref{fig:computational_efficiency} where we test for $N=16$ and $K/N \in \{1, 10, 100, 1k, 10k, 100k\}$.
Our main experiments used $K/N = 28.2k/16 = 1.76k$ for VIGOR, and $K/N = 14.4k/16 = 0.9k$ for KITTI.
In practice, memory will thus be the limiting factor for determining the number of poses that can be used.

%% file: supp-text5.tex
\section*{E. Visualization: SliceMatch Predictions}\label{sec:visual_slicematch_predictions}

Here we provide extra qualitative results for our experiments in the main paper (see Section~\ref{subsec:same-area}).
 
\begin{figure*}[b]
    \centering
    \subfloat{{\includegraphics[width=3.3cm]{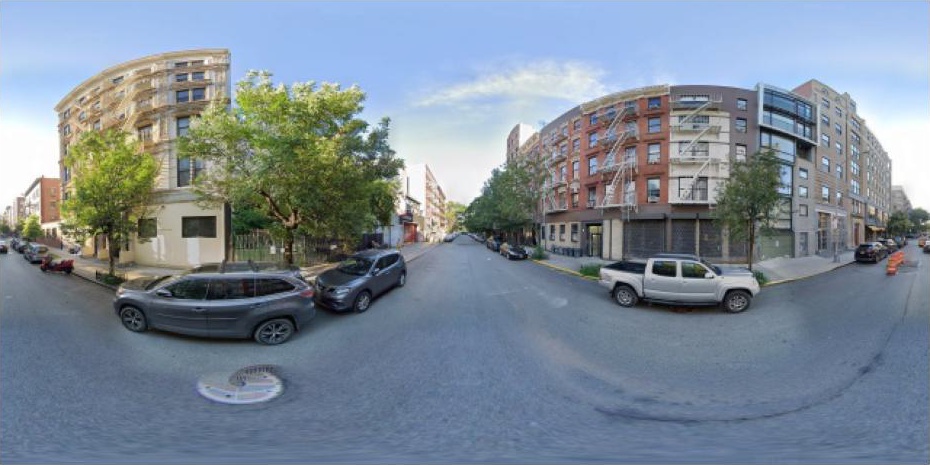}}}
    \hspace{-5mm}%
    \qquad
    \subfloat{{\includegraphics[width=3.3cm]{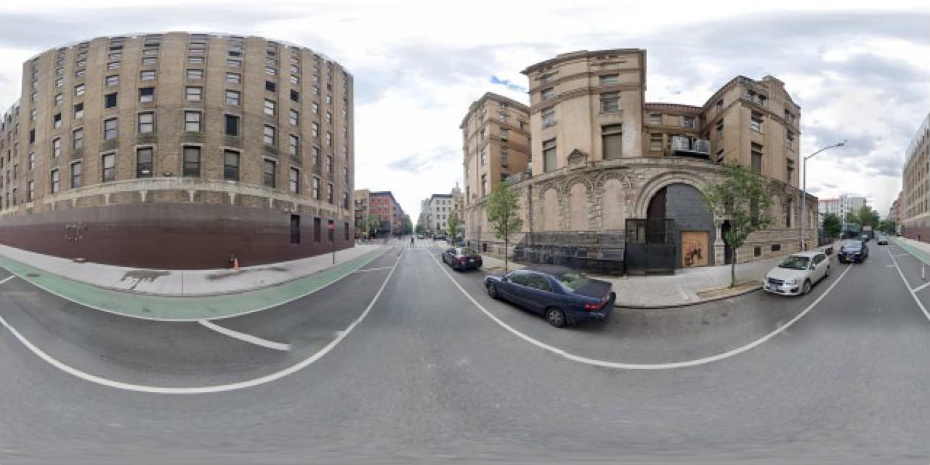}}}
    \hspace{-5mm}%
    \qquad
    \subfloat{{\includegraphics[width=3.3cm]{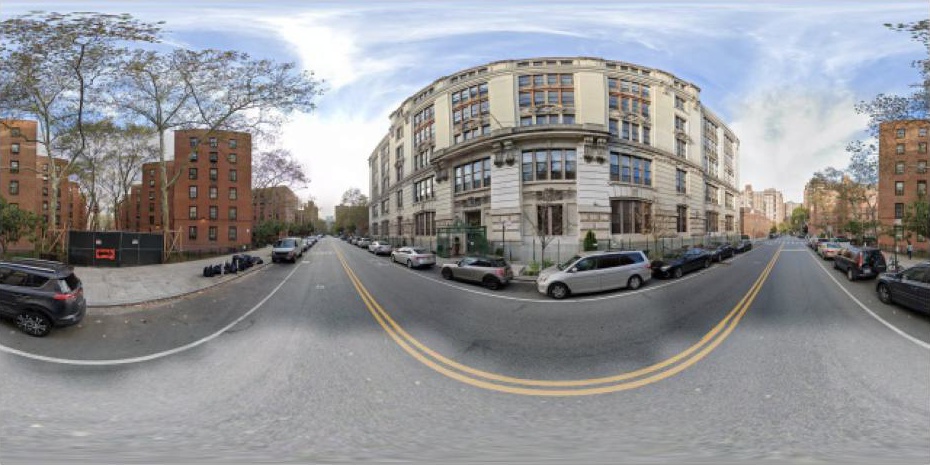}}}
    \hspace{-5mm}%
    \qquad
    \subfloat{{\includegraphics[width=3.3cm]{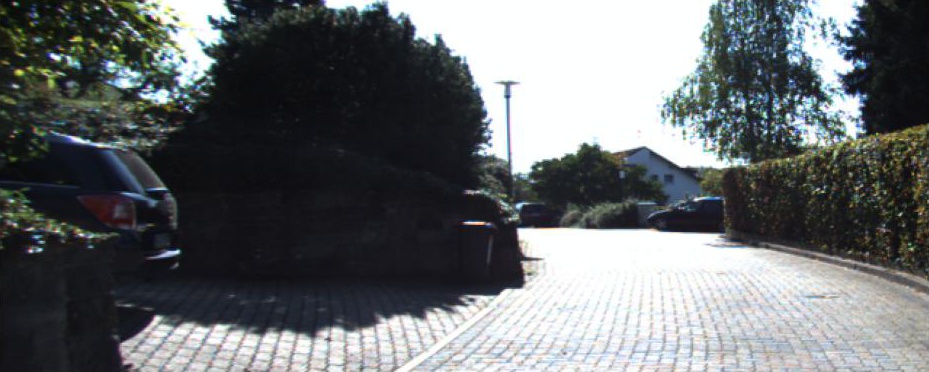}}}
    \hspace{-5mm}%
    \qquad
    \subfloat{{\includegraphics[width=3.3cm]{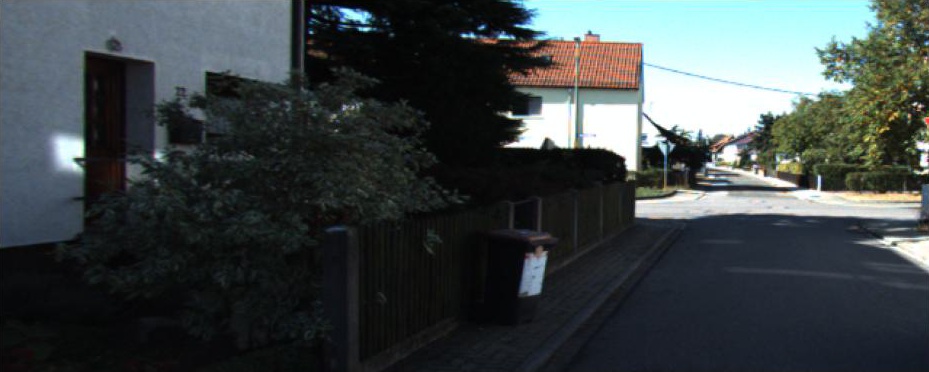}}}
    \vspace{2mm}%
    \qquad
    \subfloat[\centering VIGOR example 1]{\setcounter{subfigure}{0}%resets subfigure to (a)
    {\includegraphics[width=3.3cm]{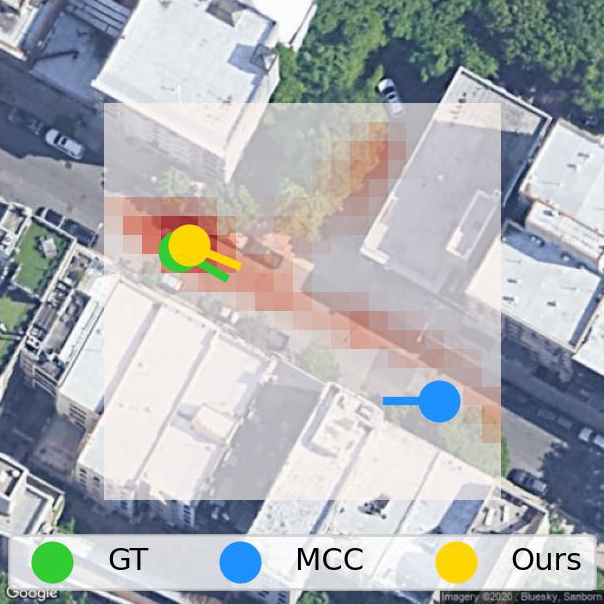}}}
    \hspace{-5mm}%
    \qquad
    \subfloat[\centering VIGOR example 2]{{\includegraphics[width=3.3cm]{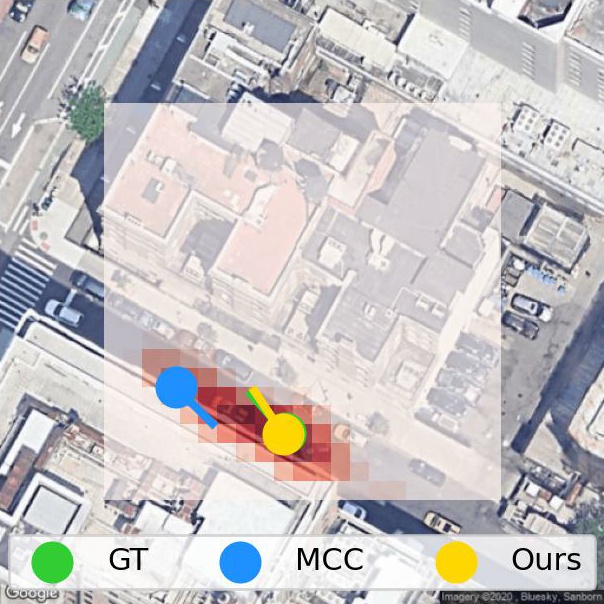}}}
    \hspace{-5mm}%
    \qquad
    \subfloat[\centering VIGOR example 3]{{\includegraphics[width=3.3cm]{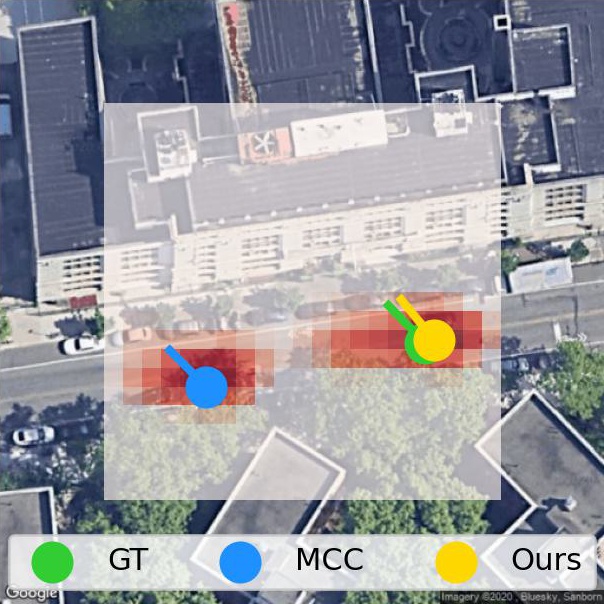}}}
    \hspace{-5mm}%
    \qquad
    \subfloat[\centering KITTI example 1]{{\includegraphics[width=3.3cm]{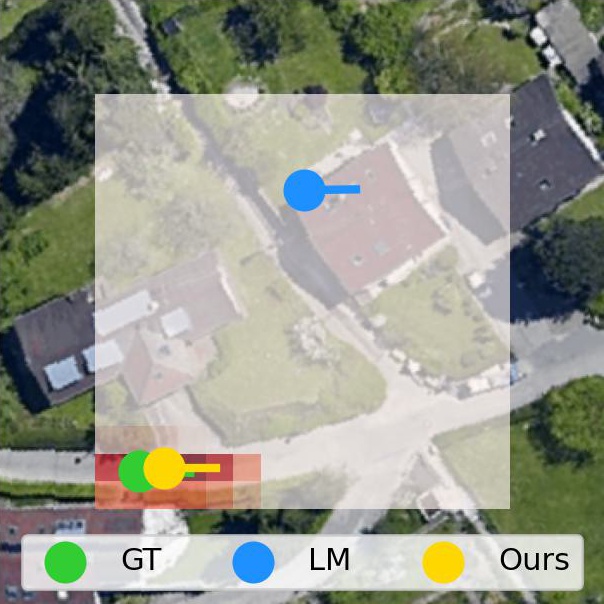}}}
    \hspace{-5mm}%
    \qquad
    \subfloat[\centering KITTI example 2]{{\includegraphics[width=3.3cm]{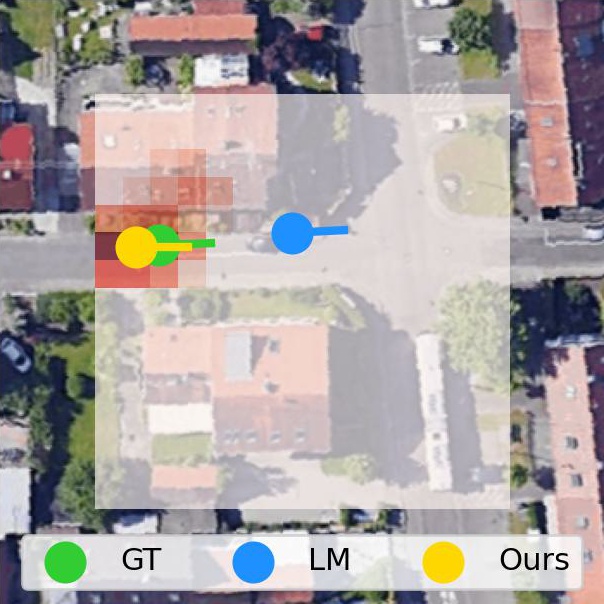}}}
    \vspace{5mm}%
    \qquad
    \subfloat{{\includegraphics[width=3.3cm]{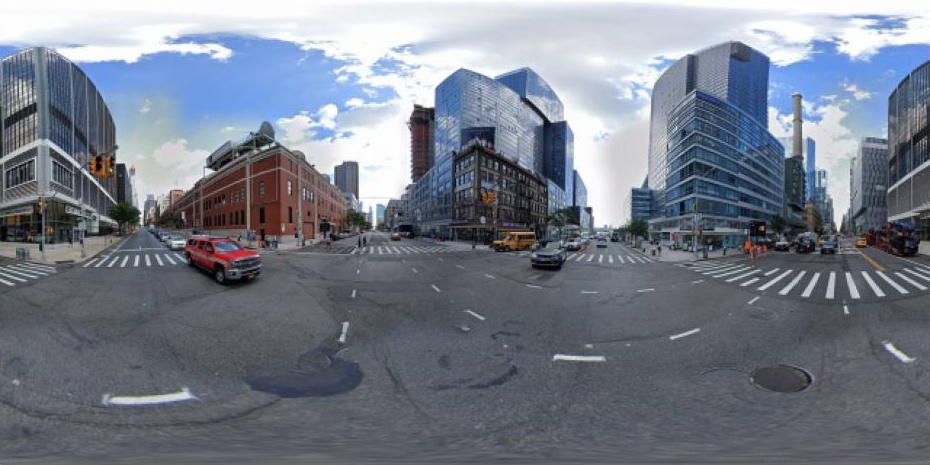}}}
    \hspace{-5mm}%
    \qquad
    \subfloat{{\includegraphics[width=3.3cm]{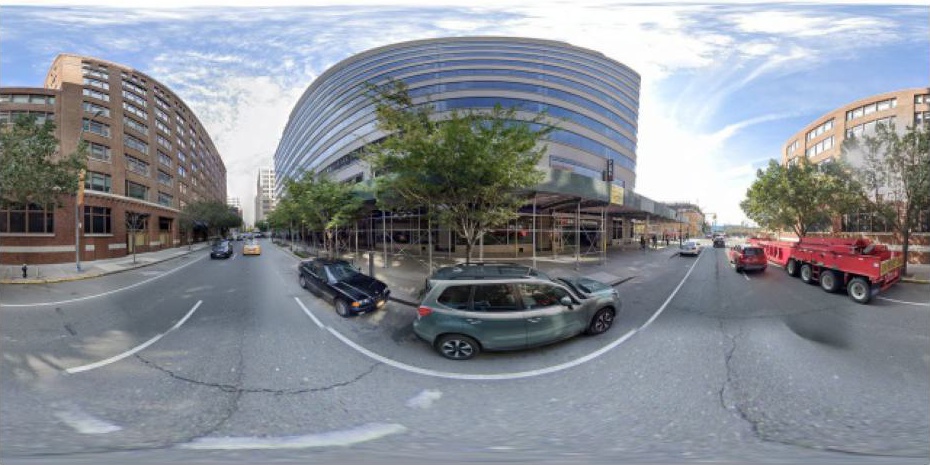}}}
    \hspace{-5mm}%
    \qquad
    \subfloat{{\includegraphics[width=3.3cm]{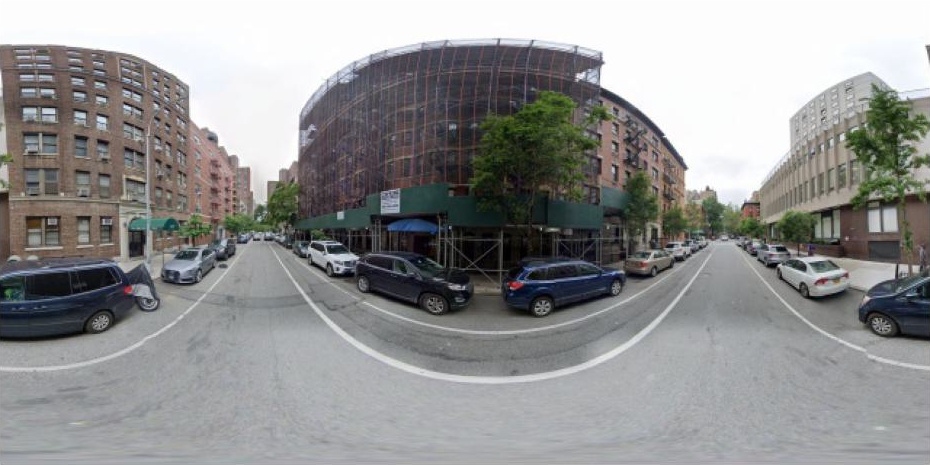}}}
    \hspace{-5mm}%
    \qquad
    \subfloat{{\includegraphics[width=3.3cm]{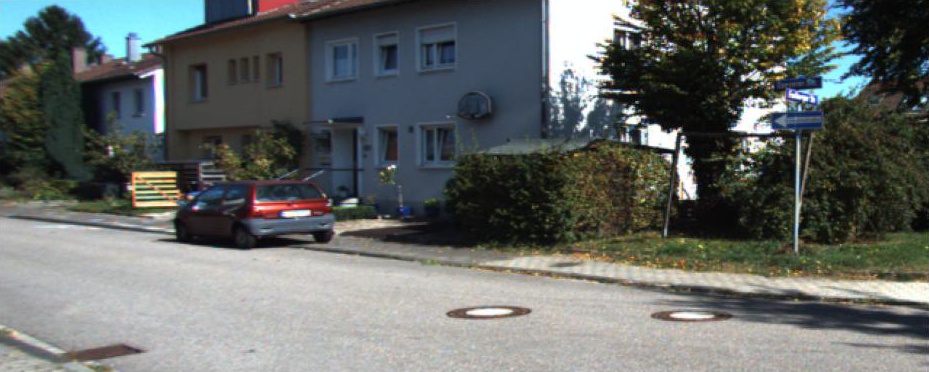}}}
    \hspace{-5mm}%
    \qquad
    \subfloat{{\includegraphics[width=3.3cm]{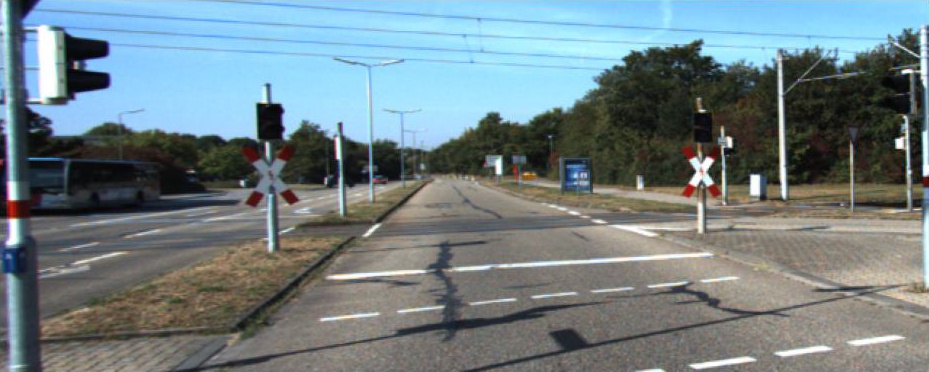}}}
    \vspace{2mm}%
    \qquad
    \subfloat[\centering VIGOR example 4]{\setcounter{subfigure}{5}%resets subfigure to (f)
    {\includegraphics[width=3.3cm]{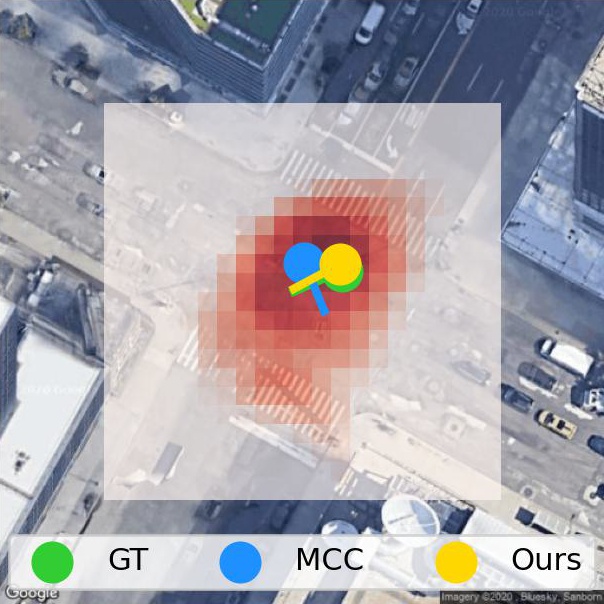}}}
    \hspace{-5mm}%
    \qquad
    \subfloat[\centering VIGOR example 5]{{\includegraphics[width=3.3cm]{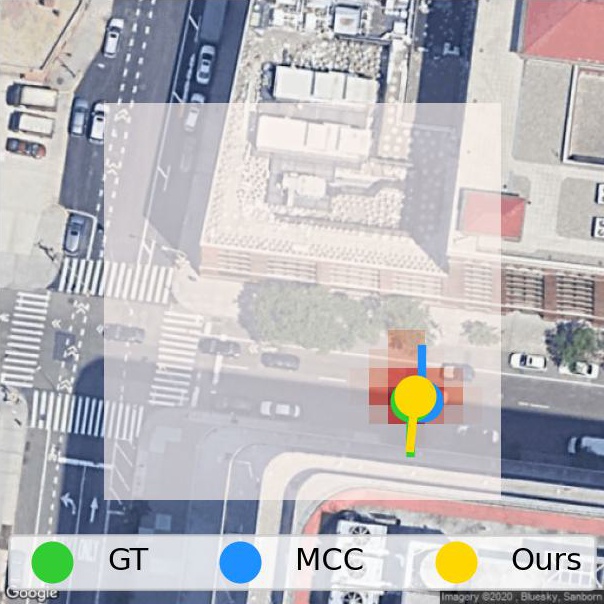}}}
    \hspace{-5mm}%
    \qquad
    \subfloat[\centering VIGOR example 6]{{\includegraphics[width=3.3cm]{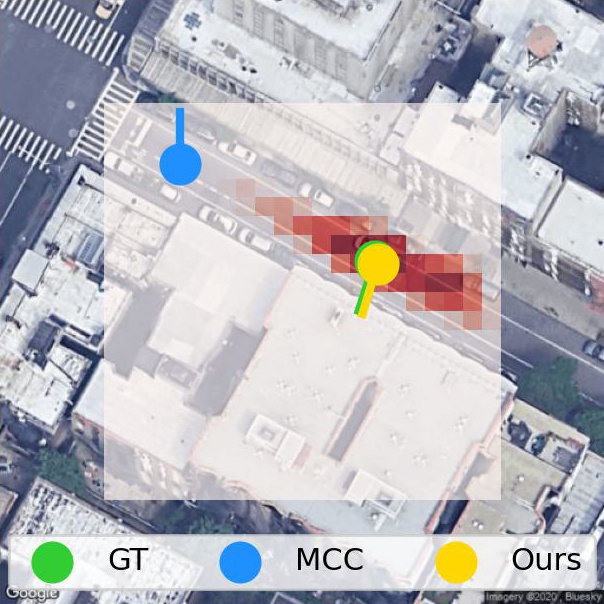}}}
    \hspace{-5mm}%
    \qquad
    \subfloat[\centering KITTI example 3]{{\includegraphics[width=3.3cm]{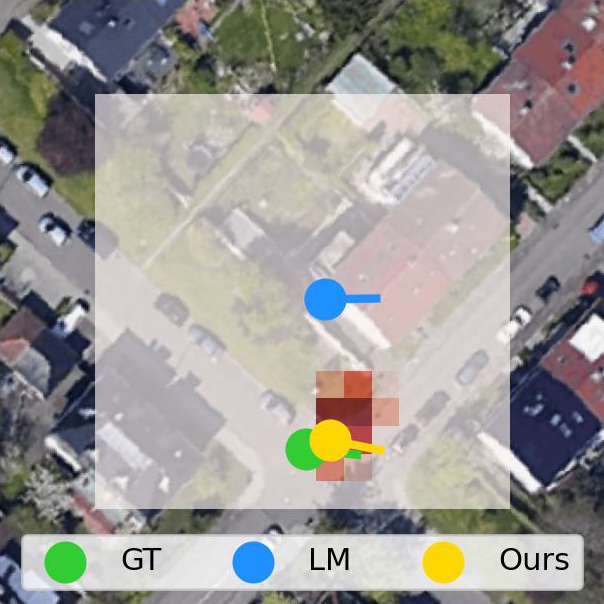}}}
    \hspace{-5mm}%
    \qquad
    \subfloat[\centering KITTI example 4]{{\includegraphics[width=3.3cm]{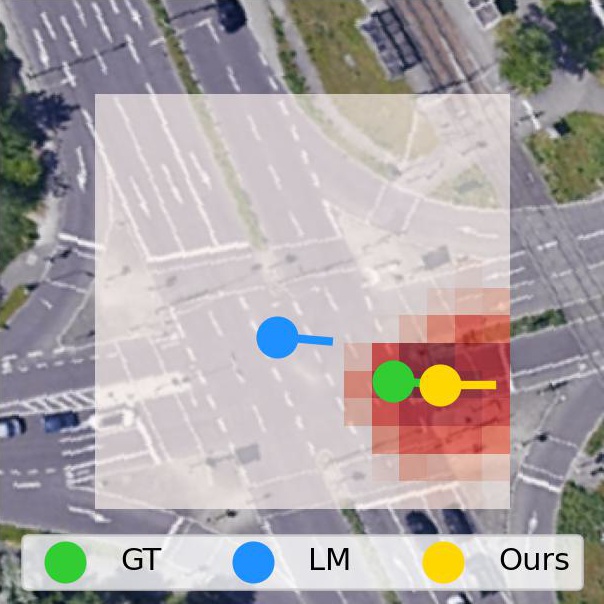}}}
    \caption{
    \textbf{Qualitative evaluation of SliceMatch on VIGOR~\cite{vigor} and KITTI~\cite{geiger2013vision, shi2022beyond}: successful cases.} Top row: input ground image. Bottom row: GT and pose estimation results overlayed on input aerial image. Red shading indicates the highest similarity score between the ground descriptor and the aerial descriptors among all orientations at that location.}
    \label{fig:slicematch_visuals_succes}
\end{figure*}

Figure~\ref{fig:slicematch_visuals_succes} shows successful predictions of SliceMatch for the VIGOR~\cite{vigor} and KITTI~\cite{geiger2013vision, shi2022beyond} datasets. In Figure~\ref{fig:slicematch_visuals_succes}a, Figure~\ref{fig:slicematch_visuals_succes}b, Figure~\ref{fig:slicematch_visuals_succes}c, and Figure~\ref{fig:slicematch_visuals_succes}h, it can be seen that the predicted similarity map is aligned with the road and that the predicted orientation is in line with the orientation of the ground image. In contrast, in Figure~\ref{fig:slicematch_visuals_succes}a, Figure~\ref{fig:slicematch_visuals_succes}b, Figure~\ref{fig:slicematch_visuals_succes}g, and Figure~\ref{fig:slicematch_visuals_succes}h, MCC~\cite{zimin_metricloc} predicts a location on the road, but the orientation sometimes differs 180 degrees from that of the ground image. In Figure~\ref{fig:slicematch_visuals_succes}d and Figure~\ref{fig:slicematch_visuals_succes}i, LM~\cite{shi2022beyond} converges to a location on the roof of a building for some image pairs.

\begin{figure*}[h]
    \centering
    \subfloat{{\includegraphics[width=3.3cm]{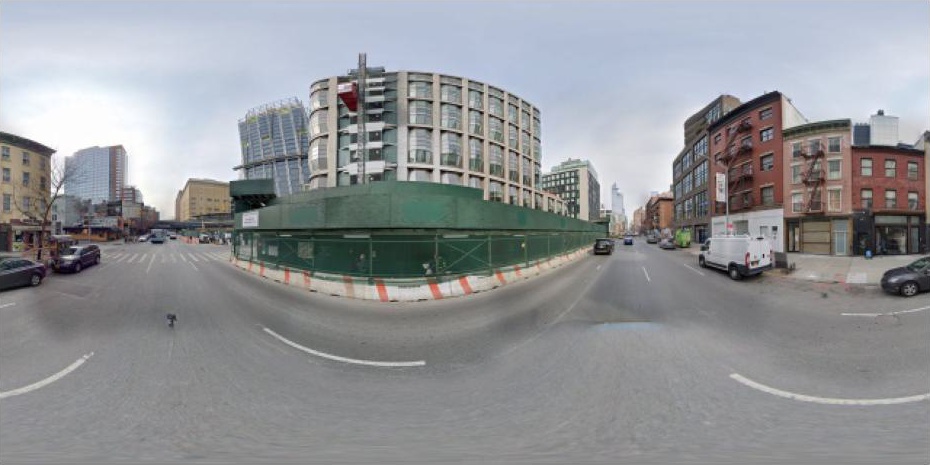}}}
    \hspace{-5mm}%
    \qquad
    \subfloat{{\includegraphics[width=3.3cm]{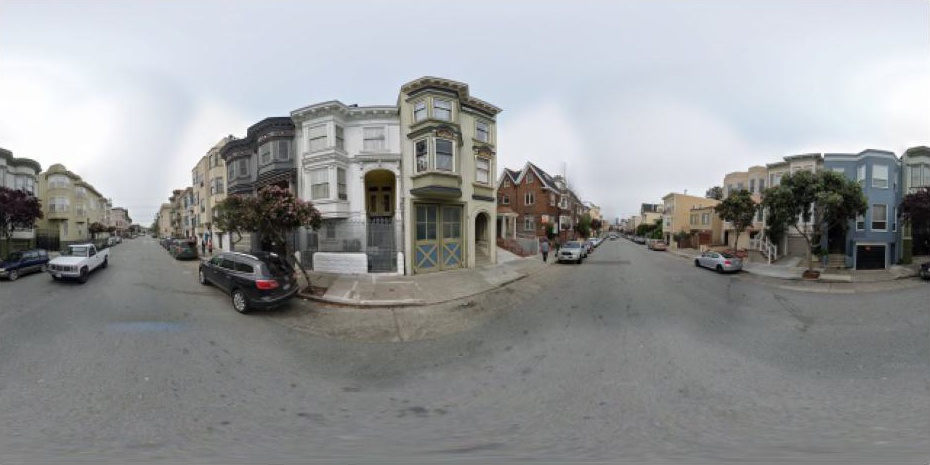}}}
    \hspace{-5mm}%
    \qquad
    \subfloat{{\includegraphics[width=3.3cm]{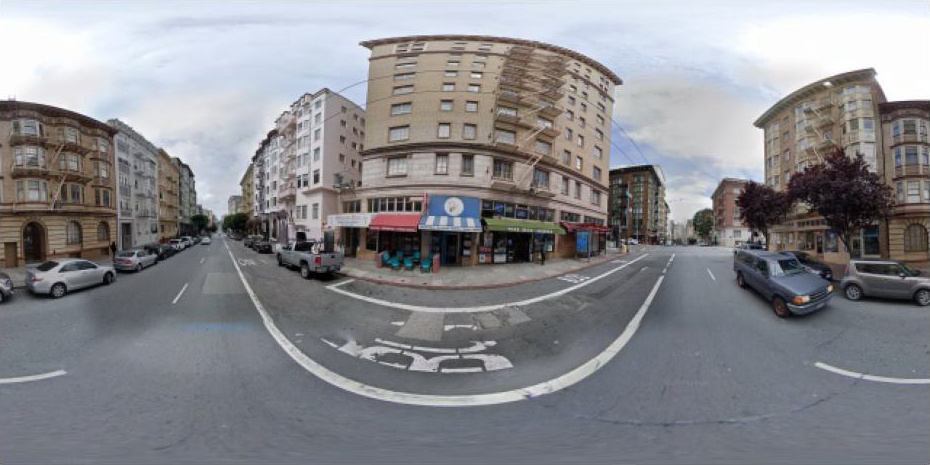}}}
    \hspace{-5mm}%
    \qquad
    \subfloat{{\includegraphics[width=3.3cm]{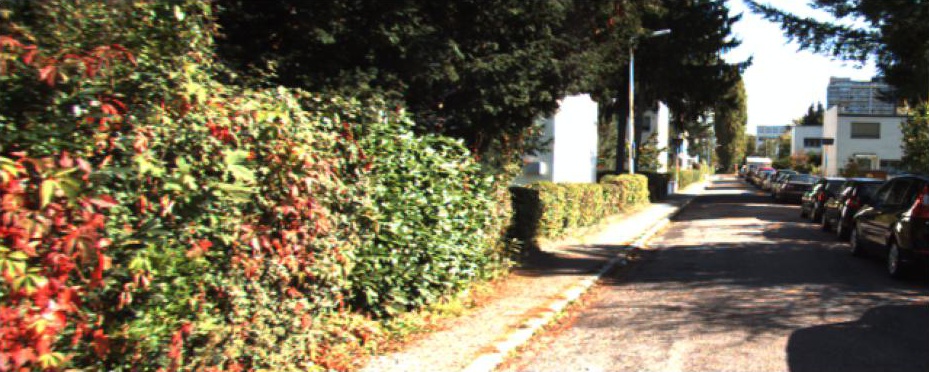}}}
    \hspace{-5mm}%
    \qquad
    \subfloat{{\includegraphics[width=3.3cm]{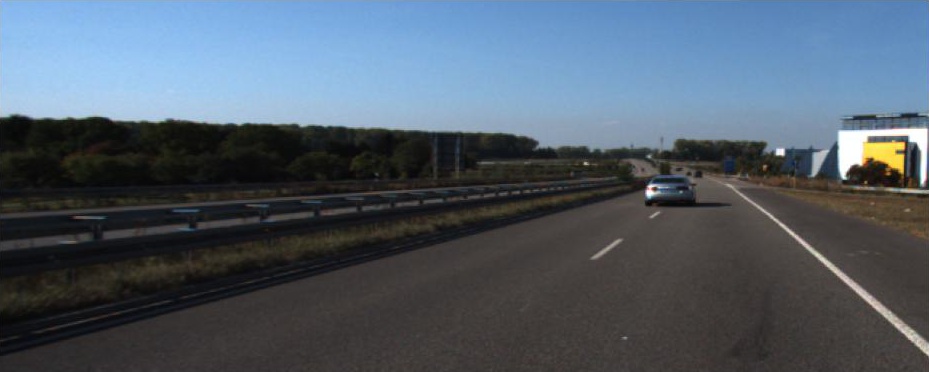}}}
    \vspace{2mm}%
    \qquad
    \subfloat[\centering VIGOR example 1]{\setcounter{subfigure}{0}%resets subfigure to (a)
    {\includegraphics[width=3.3cm]{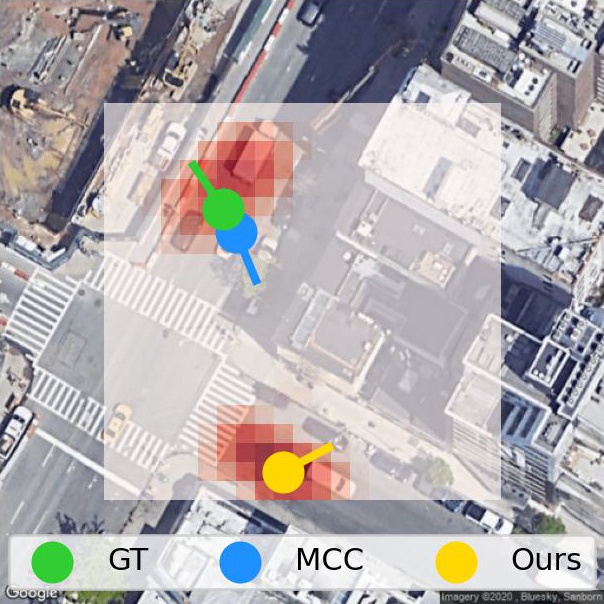}}}
    \hspace{-5mm}%
    \qquad
    \subfloat[\centering VIGOR example 2]{{\includegraphics[width=3.3cm]{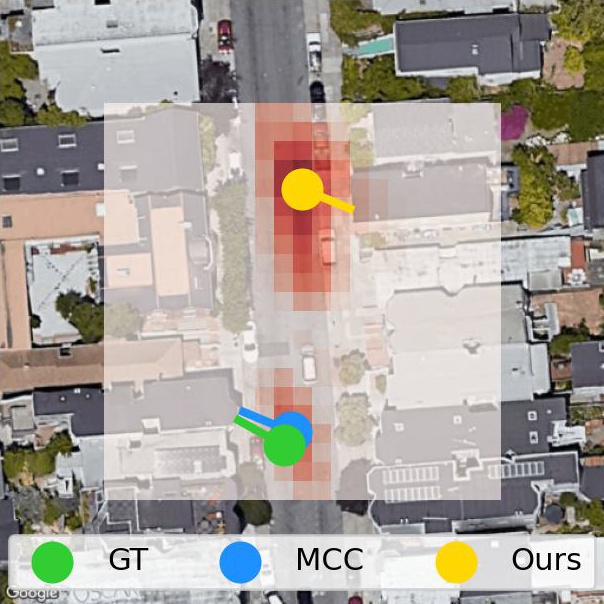}}}
    \hspace{-5mm}%
    \qquad
    \subfloat[\centering VIGOR example 3]{{\includegraphics[width=3.3cm]{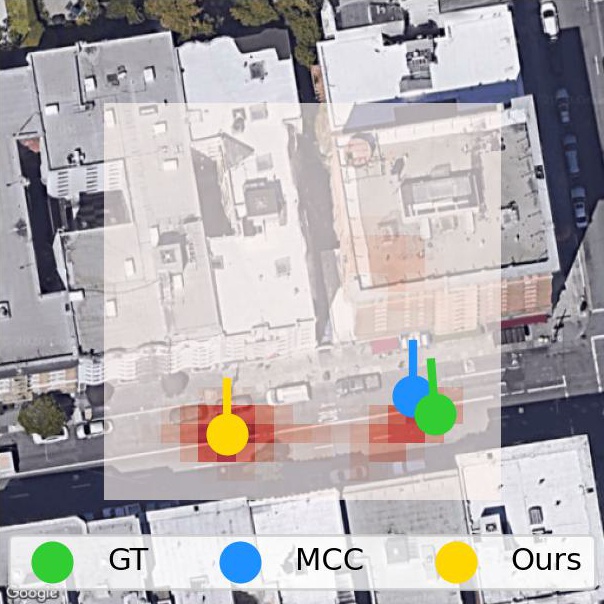}}}
    \hspace{-5mm}%
    \qquad
    \subfloat[\centering KITTI example 1]{{\includegraphics[width=3.3cm]{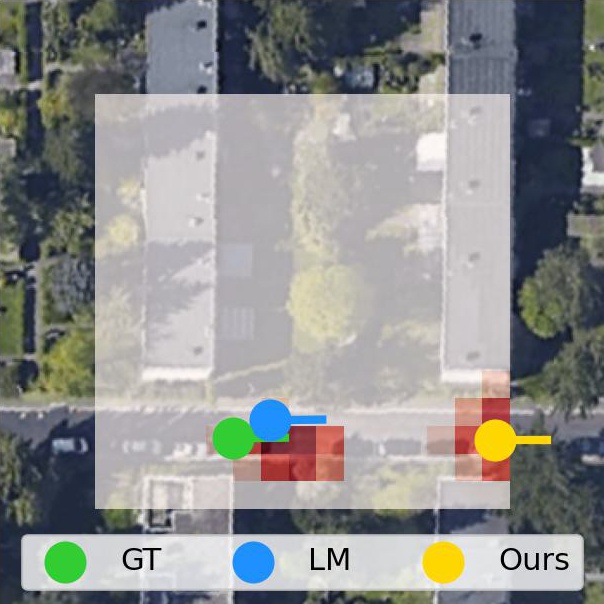}}}
    \hspace{-5mm}%
    \qquad
    \subfloat[\centering KITTI example 2]{{\includegraphics[width=3.3cm]{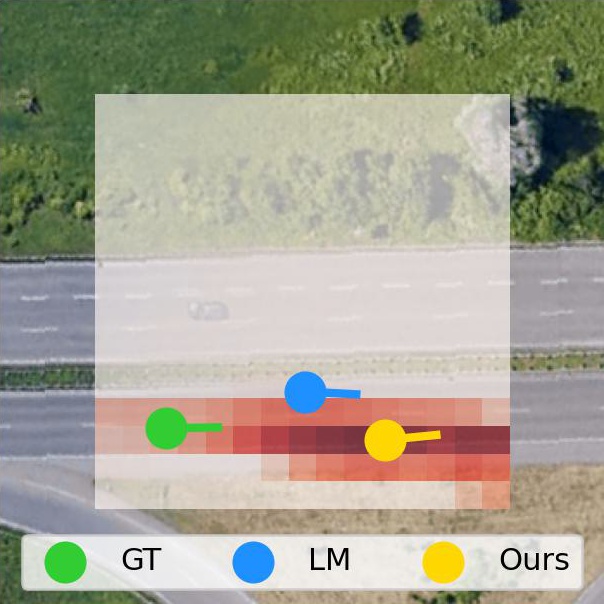}}}
    \caption{
    \textbf{Qualitative evaluation of SliceMatch on VIGOR~\cite{vigor} and KITTI~\cite{geiger2013vision, shi2022beyond}: failure cases.} Top row: input ground image. Bottom row: GT and pose estimation results overlayed on input aerial image. Red shading indicates highest similarity score between the ground descriptor and the aerial descriptors among all orientations at that location.}
    \label{fig:slicematch_visuals_failure}
\end{figure*}

Figure~\ref{fig:slicematch_visuals_failure} shows some failure cases of SliceMatch. SliceMatch can predict a multi-modal similarity map. In Figure~\ref{fig:slicematch_visuals_failure}a, Figure~\ref{fig:slicematch_visuals_failure}b, Figure~\ref{fig:slicematch_visuals_failure}c and Figure~\ref{fig:slicematch_visuals_failure}d, it can be seen that SliceMatch predicts two peaks, but the wrong peak is used as the prediction. The ground image of Figure~\ref{fig:slicematch_visuals_failure}e contains few discriminating objects and this can be observed in the predicted similarity map. SliceMatch predicts uncertainty aligned with the road. The lateral error is small, however, the longitudinal error is large.

